\documentclass[10pt,twocolumn,letterpaper]{article}

\usepackage[pagenumbers]{cvpr}
\usepackage[accsupp]{axessibility}

\usepackage{arydshln}
\usepackage{booktabs,array}
\usepackage{colortbl}
\usepackage{tabularx}
\usepackage{multirow}
\usepackage{wrapfig}
\usepackage{graphicx}
\usepackage{amsmath,amssymb,amsfonts}
\usepackage{algorithmic}
\usepackage{graphicx}
\usepackage{textcomp}
\usepackage{xcolor}
\usepackage{color}
\usepackage{xspace}
\usepackage{balance}
\usepackage[skip=8pt]{caption}
\renewcommand{\b}[1]{{\bf #1}}

\newcommand{\bb}[1]{\textbf{#1}}

\usepackage{bbding}

\newcolumntype{C}[1]{>{\centering\arraybackslash}p{#1}}
\newcolumntype{R}[1]{>{\raggedleft\arraybackslash}p{#1}}
\newcommand{\ru}{\rule{0mm}{3mm}}

\newcommand{\NAME}{{B-Free}}

\definecolor{linkblue}{rgb}{0.21,0.49,0.74}
\usepackage[pagebackref,breaklinks,colorlinks,allcolors=linkblue]{hyperref}

\title{A Bias-Free Training Paradigm for More General AI-generated Image Detection}

\author{Fabrizio Guillaro\textsuperscript{1} \ \ \ 
Giada Zingarini\textsuperscript{1} \ \ \ 
Ben Usman\textsuperscript{2} \ \ \  
Avneesh Sud\textsuperscript{2} \ \ \ \\
Davide Cozzolino\textsuperscript{1} \ \ \ 
Luisa Verdoliva\textsuperscript{1} \\[3mm]
{\textsuperscript{1}University Federico II of Naples \ \ \ \ \ \textsuperscript{2}Google DeepMind}\\
}

\begin{document}

\twocolumn[{
\maketitle
\vspace*{-0.15in}
\centering
\includegraphics[width=1.0\linewidth, page=1,clip, trim=50 340 55 0]{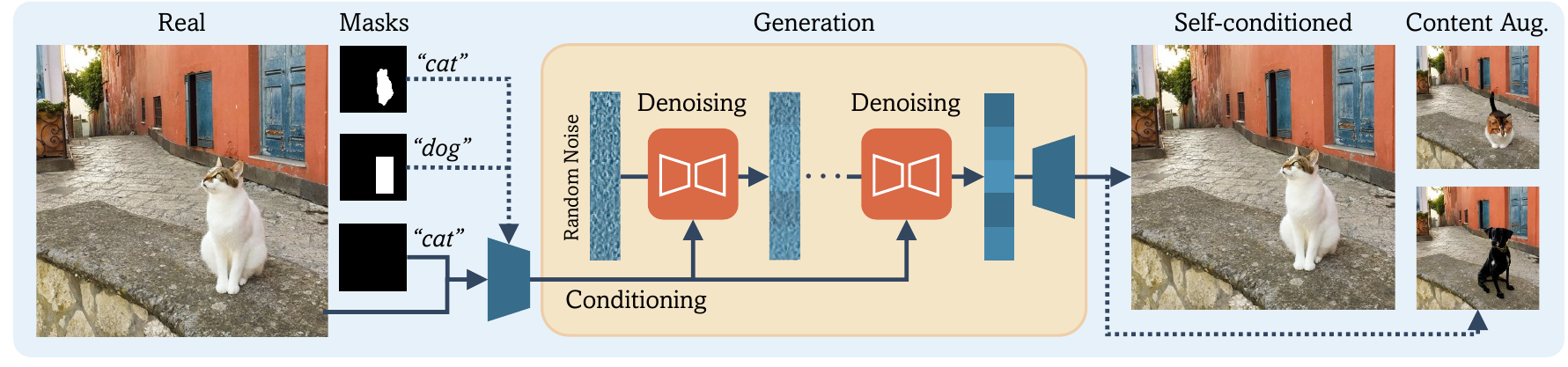}
\captionof{figure}{We introduce a new training paradigm for AI-generated image detection. To avoid possible biases, we generate synthetic images from self-conditioned reconstructions of real images and include augmentation in the form of inpainted versions. This allows to avoid semantic biases. As a consequence, we obtain better generalization to unseen models and better calibration than SoTA methods.
    }
\vspace{0.25in}
\label{fig:teaser}
}
]

\begin{abstract} 
Successful forensic detectors can produce excellent results in supervised learning benchmarks but struggle to transfer to real-world applications. We believe this limitation is largely due to inadequate training data quality. While most research focuses on developing new algorithms, less attention is given to training data selection, despite evidence that performance can be strongly impacted by spurious correlations such as content, format, or resolution. A well-designed forensic detector should detect generator specific artifacts rather than reflect data biases. To this end, we propose B-Free, a bias-free training paradigm, where fake images are generated from real ones using the conditioning procedure of stable diffusion models. This ensures semantic alignment between real and fake images, allowing any differences to stem solely from the subtle artifacts introduced by AI generation. Through content-based augmentation, we show significant improvements in both generalization and robustness over state-of-the-art detectors and more calibrated results across 27 different generative models, including recent releases, like FLUX and Stable Diffusion 3.5. Our findings emphasize the importance of a careful dataset design, highlighting the need for further research on this topic.
Code and data are publicly available at \url{https://grip-unina.github.io/B-Free/}. 
\end{abstract}

\section{Introduction}
\label{sec:intro}

The rise of generative AI has revolutionized the creation of synthetic content, 
enabling easy production of high-quality sophisticated media, even for individuals without deep technical expertise. Thanks to user-friendly interfaces and pretrained models, users can create synthetic content such as text, images, music, and videos through simple inputs or prompts \cite{Zhan2023multimodal}. 
This accessibility has democratized content creation, enabling professionals in fields like design, marketing, and entertainment to leverage AI for creative purposes. However, this raises concerns about potential misuse, such as the creation of deepfakes, misinformation, and challenges related to intellectual property and content authenticity \cite{Epstein2023art, Barrett2024identifying, Lin2024detecting}.

\begin{figure*}[t!]
    \centering
    \includegraphics[page=1, width=1\linewidth, clip, trim=0 0 0 0]{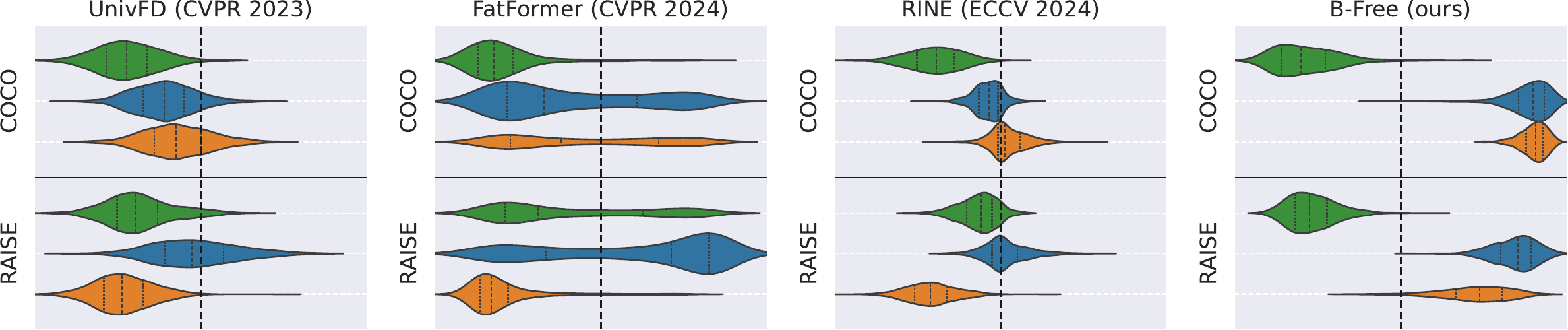} \\
    \includegraphics[page=1, width=0.4\linewidth, clip, trim=0 0 0 0]{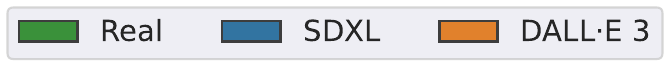}
    \caption{\small Forensic detectors can exhibit opposite behaviors depending on their training dataset. The four plots show the prediction distributions for three ViT-based detectors, UnivFD \cite{Ojha2023towards}, FatFormer \cite{Liu2024forgery} and RINE \cite{Koutlis2024leveraging}, and the proposed one. The fake images (SD-XL or DALL-E 3) are generated from images of a single dataset (RAISE on top, COCO on the bottom) and tested only against real images of the same dataset (Synthbuster \cite{Bammey2023synthbuster} and the test dataset from \cite{Cozzolino2024raising}). We observe that for the same detector (e.g., RINE) and the same fake-image generator (e.g., DALL-E 3) the score distributions can vary significantly depending on the dataset used, going from real (left of the dotted line) to fake (right of the dotted line) or vice versa. This is likely due to the presence of biases in the training set that heavily impact the detector prediction. Our detector, on the other hand, shows consistent and correct results.}
    \label{fig:biases}
\end{figure*}

Key challenges for current GenAI image detectors include generalization — detecting synthetic generators not present in the training set — and ensuring robustness against image impairments caused by online sharing, such as compression, resizing, and cropping \cite{Tariang2024synthetic}. In this context, large pre-trained vision-language models like CLIP \cite{Radford2021learning} have demonstrated impressive resilience to these distribution shifts \cite{Ojha2023towards}. The success of these models in forensic applications suggests that pre-training on large and diverse datasets may be a promising path forward. 
An important aspect often overlooked in the current literature is the selection of good datasets to train or fine-tune such models that primarily rely on hidden, unknown signatures of generative models \cite{Marra2019DoGAN, Yu2019attributing}.
Indeed, it is important to guarantee that the detector decisions are truly based on generation-specific artifacts and not on possible dataset biases \cite{Torralba2011unbiased, Liu2024decade, Cazenavette2024fakeinversion}. In fact, datasets used during the training and testing phases of forensic classifiers could be affected by different types of polarization. 

Format issues have been the Achilles' heel of forensic detectors since at least 2013,
when \cite{Cattaneo2014possible} recognized that a dataset for image tampering detection \cite{Dong2013casia} included
forged and pristine images compressed with different JPEG quality factors. Therefore, a classifier trained to discrminate tampered and pristine images 
may instead learn their different processing histories. This issue has been 
highlighted in \cite{Grommelt2024fake} with reference to datasets of synthetic and real images. In fact, the former are often created in a lossless format (PNG), while the latter are typically compressed in lossy formats like  JPEG. Again a classifier could learn coding inconsistencies instead of forensic clues. Likewise it could learn resampling artifacts, as it was recently shown in \cite{Rajan2024effectiveness} - in this case 
a bias was introduced by resizing all the real images from the LAION dataset to the same resolution, while keeping the fake ones unaltered. 

Forensic clues are subtle and often imperceptible to the human eye, making it easy to introduce biases when constructing the training and test sets, as well as the evaluation protocol. Semantic content itself can also represent a source of bias. For this reason, several recent proposals \cite{Bammey2023synthbuster, Cozzolino2024raising, Baraldi2024contrastive} take great care to include pairs of real and fake images characterized by the same prompts when building a training or test dataset. To gain better insights about the above issues, in Fig.~\ref{fig:biases} we show the performance of three SoTA ViT-based approaches  \cite{Ojha2023towards, Koutlis2024leveraging, Liu2024forgery} in distinguishing real images from fake images generated by SD-XL and DALL-E 3. For each method we consider two settings: in the first case, real images come from the RAISE dataset \cite{Nguyen2015raise} and fakes are generated starting from images of the same dataset. The second case uses COCO as source of reals instead of RAISE. We note an inconsistent behavior of SOTA forensic detectors on the same synthetic generator which can be caused by the presence of biases during training. 
FakeInversion \cite{Cazenavette2024fakeinversion} proposes an effective approach towards semantic alignment of training data using reverse image search to find matching reals, but fails to capture real image distribution after 2021. 

To mitigate potential dataset biases, in this work we propose a new training paradigm, \NAME, where synthetic images are generated using self-conditioned reconstructions of real images and incorporate augmented, inpainted variations. This approach helps to prevent semantic bias and potential misalignment in coding formats.
Furthermore, the model avoids resizing operations that can be particularly harmful by washing out the subtle low-level forensic clues \cite{Gragnaniello2021GAN}.  
Overall, we make the following contributions:
\begin{itemize}
\item We propose a large curated training dataset of 51k real and 309k fake images. Real images are sourced from COCO, while synthetic images are self-conditioned generations using Stable Diffusion~2.1. This helps the detector to focus on artifacts related to the synthetic generation process avoiding content and coding related biases.
\item We show that incorporating proper content-based augmentation leads to better-calibrated results. This ensures that in-lab performance more closely aligns with the expected performance on real-world images shared across social networks.
\item We study the effect of different distribution shifts and show that by leveraging a pre-trained large model fine-tuned end-to-end on our dataset, we achieve a SoTA accuracy superior to 90\% even on unseen generators. 
\end{itemize}

\begin{figure*}[t!]
    \centering
    \includegraphics[width=0.98\linewidth, page=2,clip, trim=10 310 10 0]{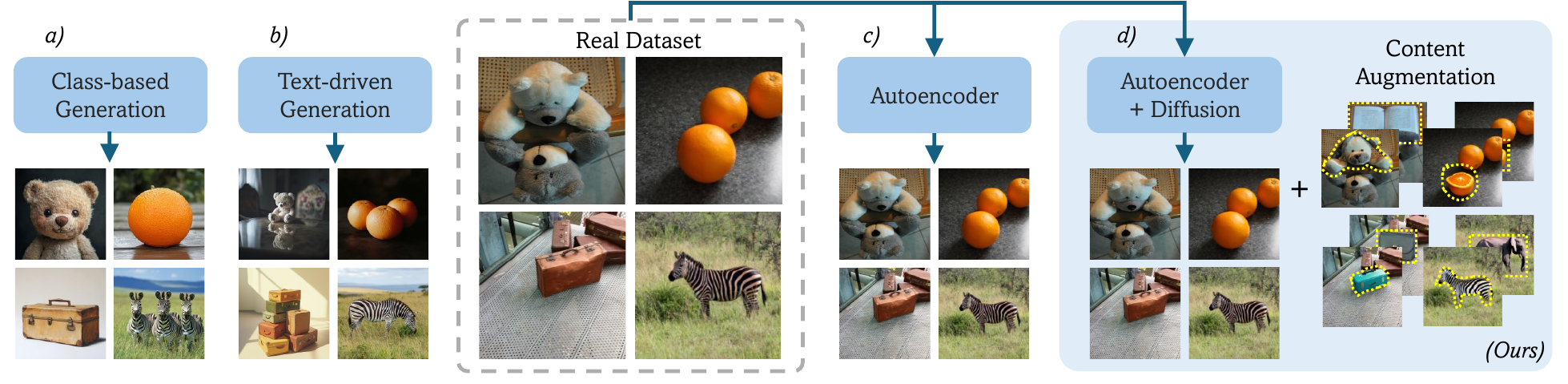}
 
     \caption{Overview of existing ($a$, $b$, $c$) and proposed ($d$) strategies for building an aligned training dataset. Some methods try to match synthetic images to the corresponding real images by using class-based generation ($a$) or text-to-image generation with real images' descriptions ($b$). In ($c$) real images are fed to an autoencoder to generate a \textbf{reconstructed fake} with the same content. Differently from ($c$), in our approach  a \textbf{self-conditioned fake} is generated using diffusion steps ($d$), and we also add a content augmentation step.}
    \label{fig:training_approaches}
\end{figure*}

\section{Related Work} 
\label{sec:related}

A well-curated training set is of vital importance for any data-driven method. 
In recent years this awareness has much grown also in the forensic field 
and there have been many efforts in this direction following two main lines of work: 
{\em i)}  forming a reliable dataset by carefully selecting ``natural'' fakes, or
{\em ii)} creating a fully synthetic dataset by injecting forensic artifacts in real images.

\vspace{2mm}
\noindent
\bb{Selecting good natural training data.}
Wang et al.'s paper \cite{Wang2020cnn} was among the first to demonstrate the importance of selecting a suitable training set for gaining generalization to unseen synthetic generators. The selected dataset included images from a single generation architecture (ProGAN) and 20 different real/false categories (Fig.~\ref{fig:training_approaches}.a) and included augmentation in the form of common image post-processing operations, such as blurring and compression. Results clearly show that generalization and robustness strongly benefit from the many different categories included during training as well as from the augmentation procedure. In fact, this dataset has been widely utilized in the literature, where researchers follow a standard protocol assuming the knowledge of one single generative model during training. 
This scenario describes a typical real world situation where new generative architectures are unknown at test time. 

The dataset proposed in \cite{Wang2020cnn} was used in \cite{Ojha2023towards} to fine-tune a CLIP model with a single learnable linear layer, achieving excellent generalization not only on GAN models but also on Diffusion-based synthetic generators never seen during training. Likewise, it was used in \cite{Koutlis2024leveraging} to train a CNN classifier that leverages features extracted from CLIP's intermediate layers to better exploit low-level forensic features. In \cite{Sha2023defake, Tan2024c2p} image captions (either paired to the dataset images or generated from them) were used as additional input for a joint analysis during training.
The approach proposed in \cite{Liu2024forgery} is trained using only 4 classes out of the 20 categories proposed in \cite{Wang2020cnn}, as well as other recent methods
\cite{Tan2024rethinking, Tan2023learning, Tan2024c2p}.

Alternatively, some methods rely on datasets comprising images from a single diffusion-based generator, such as Latent Diffusion \cite{Corvi2023detection, Cozzolino2024raising, Cazenavette2024fakeinversion}, Guided Diffusion \cite{Wang2023dire} or Stable Diffusion \cite{Sha2023defake, Lanzino2024faster}.
Prior work \cite{Cozzolino2024raising, Cazenavette2024fakeinversion} highlights the importance of aligning both training and test data in terms of semantic content.
This choice allowed to better exploit the potential of fixed-pretraining CLIP features by strongly reducing the number of images needed for fine-tuning \cite{Cozzolino2024raising}. 
In addition, it has the key merit of reducing the dataset content bias, thus allowing for better quality training, and is also adopted in other approaches both during training \cite{Amoroso2023parents, Baraldi2024contrastive} and at test time to carry out a fairer evaluation \cite{Bammey2023synthbuster}.

\vspace{2mm}
\noindent
\bb{Creating training data by artifact injection.}
A different line of research is to create simulated fake images by injecting traces of the generative process in real images. A seminal work along this line was done by Zhang et al. \cite{Zhang2019detecting} for GAN image detection. The idea is to simulate artifacts shared by several generators. These peculiar traces are caused by the up-sampling processes included in the generation pipeline and show up as peaks in the frequency domain. Besides these frequency peaks, synthetic images, both GAN-based and diffusion-based, have been shown to exhibit spectral features that are very different from those of natural images \cite{Durall2020watch, Dzanic2020Fourier}. In fact, real images exhibit much richer spectral content at intermediate frequencies than synthetic ones \cite{Yang2022diffusion, Corvi2023intriguing}.

For GAN-generated images, producing realistic simulated fakes requires training the generation architecture specifically for this task \cite{Zhang2019detecting, Jeong2022fingerprintNet}. In contrast, diffusion-based image generation can leverage a pre-trained autoencoder embedded within the generation pipeline, which projects images into a latent space without the need for additional training \cite{Cozzolino2024raising, Mandelli2024when}. This procedure has been very recently used in a concurrent work \cite{Rajan2024effectiveness} to reduce semantic biases during training (Fig.~\ref{fig:training_approaches}.c). 
Different from \cite{Rajan2024effectiveness} we generate synthetic data by also performing the diffusion steps. Later in this work we will show that this choice allows us to exploit even subtler inconsistencies at lower frequencies, enhancing the detector performance (Fig.~\ref{fig:training_approaches}.d).

\renewcommand{\b}[1]{{\bf #1}}
\newcommand{\twr}[1]{\multirow{2}{*}{\shortstack[c]{#1}}}
\newcommand{\thr}[1]{\multirow{3}{*}{\shortstack[c]{#1}}}
\newcommand{\cm}[0]{\checkmark}
\newcommand{\rob}[1]{\rotatebox{90}{#1}}

\begin{table}[t!]
    \centering
    \renewcommand{\arraystretch}{1.2} 
    \resizebox{1.0\linewidth}{!}{\small
    \begin{tabular}{rcp{1.9cm}c}
    \toprule
     \textbf{Reference} & \textbf{\# Real/ \# Fake} & \textbf{Real Source} & \textbf{\# Models} \\
    \midrule
    Synthbuster \cite{Bammey2023synthbuster} & 1k / 9k      &  RAISE     & 9 \\ 
    GenImage \cite{Zhu2024genimage}             & 1.3M / 1.3M    & ImageNet    & 8 \\  
    FakeInversion \cite{Cazenavette2024fakeinversion} & 44.7k / 44.7k  & Internet  & 13 \\ 
    SynthWildX \cite{Cozzolino2024raising}   & 500 / 1.5k  & X  &  3 \\ 
    WildRF    \cite{Cavia2024real}           & 1.25k / 1.25k &  Reddit, FB, X & unknown \\ 
    \bottomrule
    \end{tabular}
    }
    \caption{This table provides an overview of the datasets used in our evaluation, including the number of real and fake images, the sources of the real data, and the number of generative models used to create the synthetic images.
    }
    \label{tab:datasets-eval}
\end{table}

\section{Evaluation Protocol}
\label{sec:evaluation-protocol}

\subsection{Datasets}
In our experimental analysis, we want to avoid or at least minimize the influence of any possible afore-mentioned biases. To this end, we carefully select the evaluation datasets as outlined below. Experiments on further datasets are provided in the supplementary material.

\bb{To avoid format bias}, we use Synthbuster \cite{Bammey2023synthbuster}, where both real and generated images are saved in raw format. Therefore, a good performance on this dataset cannot come from the exploitation of JPEG artifacts. 
A complementary strategy to avoid format biases is to reduce the mismatch between real (compressed) and synthetic (uncompressed) images by compressing the latter.
To this end, we modified the fake class in GenImage \cite{Zhu2024genimage} by compressing images at a JPEG quality close to those used for the real class, as suggested in \cite{Grommelt2024fake}. This modified dataset, referred to as GenImage unbiased, 
comprises 5k real and 5k fake images, a small fraction of the original dataset.

\bb{To avoid content bias}, we also evaluate performance on datasets where fakes are generated using automated descriptions of real images.
In studies like \cite{Bammey2023synthbuster, Baraldi2024contrastive} these descriptions are refined into manually created prompts for text-based generation.
As a result, the generated images closely align with the content of the real images, minimizing possible biases due to semantic differences.
A more refined dataset in this regard is FakeInversion \cite{Cazenavette2024fakeinversion},
where real images are retrieved from the web using reverse image search, thus ensuring stylistic and thematic alignment with the fakes. 

\bb{To allow in-the-wild analysis}, we experiment also on datasets of real/fake images collected from the web, such as WildRF \cite{Cavia2024real} and SynthWildX \cite{Cozzolino2024raising}. Both datasets comprise images coming from several popular social networks. Tags were used to find fake images on Reddit, Facebook and X.
A short summary of all the datasets used in our evaluation is listed in Table \ref{tab:datasets-eval}.

\subsection{Metrics}
Most work on GenAI image detection measure performance by means of threshold-independent metrics, such as Area Under the Curve (AUC) or average precision (AP). These metrics indicate ideal classification performance, however the optimal separating threshold is not known and, quite often, the balanced accuracy at a fixed threshold (e.g. $0.5$) remains low, especially when there are significant differences between training and testing distributions \cite{Tariang2024synthetic}. Some papers address this problem by adjusting the threshold through a calibration procedure, assuming access to a few images from the synthetic generator under evaluation \cite{Wang2020cnn, Corvi2023detection, Ojha2023towards}. 
In a realistic situation the availability of such calibration images cannot be guaranteed.

In this work, to provide a comprehensive assessment of performance,  
we use both AUC and Accuracy at $0.5$, in addition we
compute the Expected Calibration Error (ECE) and the Negative Log-Likelihood (NLL).
ECE measures the ability of a model to provide prediction probabilities well aligned with the true probabilities. More precisely, we use the Binary ECE, which is the weighted average of the differences between the actual probability and the predicted probability across different bins \cite{Naeini2015obtaining}. 
Then, we use the balanced Negative Log-Likelihood \cite{Quinonero2006evaluating}, which evaluates the similarity between the distribution of the model's predictions and the actual data distribution, penalizing both low confidence in the correct class and overconfidence in incorrect ones. 
More details on these metrics can be found in the supplementary material.

\begin{figure}[t!]
    \centering
    \includegraphics[width=1.0\linewidth, page=3, clip, trim=150 20 130 0]{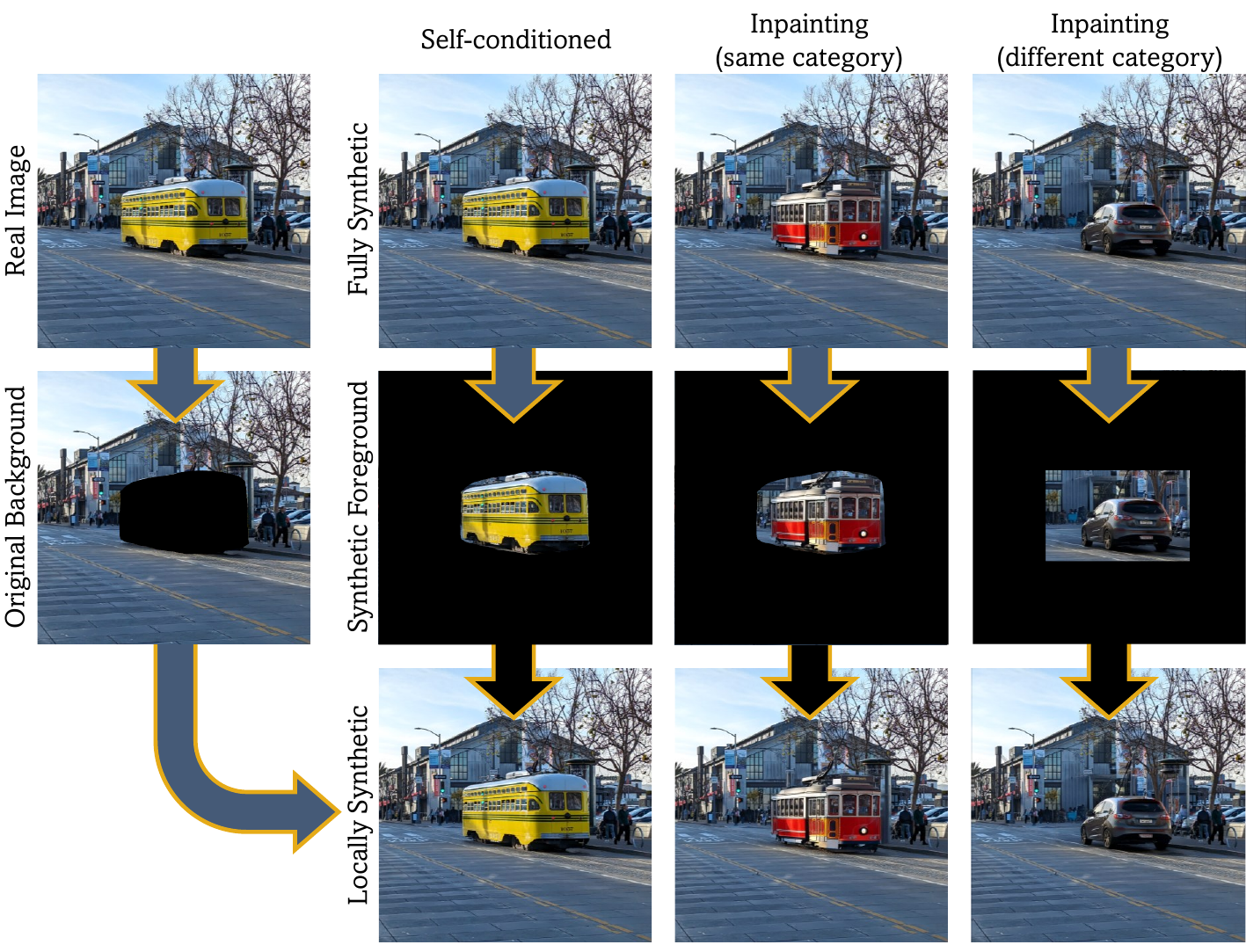}
 
     \caption{\b{Content augmentation} process. Starting with a real image, we use its generated variants (first row) and their locally manipulated versions (last row), created by replacing the original background. When inpainting with a different category, we use a bounding box instead of an object mask to allow space for new objects of varying shapes and sizes.}
    \label{fig:augmentation}
\end{figure}

\begin{table*}[t!]
    \resizebox{1.\linewidth}{!}{\small
    \begin{tabular}{cll C{0mm} C{14mm}C{14mm}C{14mm}C{14mm}C{14mm} C{14mm}C{14mm} C{14mm}C{14mm}C{14mm} C{22mm}}
    \toprule
    &  & && \multicolumn{5}{c}{\b{Synthbuster}} & \multicolumn{2}{c}{\b{New Generators}} & \multicolumn{3}{c}{\b{WildRF}} & \b{AVG} \\
    \cmidrule(lr){1-3} \cmidrule(r){5-9} \cmidrule(r){10-11} \cmidrule(r){12-14} \cmidrule(lr){15-15}
    & Method & Training setting && Midjourney  &  SDXL   & DALL·E 2  &  DALL·E 3  &  Firefly  &  FLUX & SD 3.5 & Facebook & Reddit & Twitter & AUC$\uparrow$/bAcc$\uparrow$  \\ 
    \cmidrule(lr){1-3} \cmidrule(r){5-9} \cmidrule(r){10-11} \cmidrule(r){12-14} \cmidrule(lr){15-15}
\multirow{8}{*}{\rotatebox[origin=c]{90}{AUC / bAcc}}
& \ru  paired by text    & $-$           &&    96.9  /    56.9  & \b{99.5} /    78.1  &    78.7  /    50.1  &    98.8  /    56.6  &    91.3  /    51.1  &    94.6  /    51.5  &    96.6  /    66.9  &    97.8  /    72.5  &    84.4  /    67.2  &    96.1  /    68.1  &    93.5  /    61.9  \\
& \ru  reconstructed     & $-$           && \b{100.} / \b{99.8} & \b{100.} / \b{100.} &    81.1  /    52.2  & \b{99.1} /    75.8  &    96.7  /    62.1  &    98.2  /    64.0  & \b{99.7} /    88.8  & \b{98.9} / \b{97.5} &    77.8  /    75.5  &    94.8  /    91.1  &    94.6  /    80.7  \\
& \ru  reconstructed     & inpainted     && \b{100.} / \b{99.9} & \b{100.} / \b{99.9} &    90.1  /    62.0  & \b{99.2} /    79.5  &    98.1  /    77.2  &    95.8  /    69.1  & \b{99.4} /    89.1  & \b{98.8} /    95.9  &    78.7  /    76.1  &    94.9  /    90.3  &    95.5  /    83.9  \\
& \ru  self-conditioned  & $-$           && \b{99.9} /    97.2  & \b{100.} / \b{99.2} &    90.4  /    58.1  &    98.9  /    76.1  & \b{99.4} /    89.7  &    95.4  /    59.4  & \b{100.} / \b{98.4} &    95.4  /    86.6  &    75.7  /    66.3  &    91.4  /    82.7  &    94.7  /    81.4  \\
\cdashline{2-3} \cdashline{5-15}
& \ru  self-conditioned  & cutmix/mixup  && \b{99.9} /    94.3  & \b{99.9} /    97.8  &    93.5  /    53.2  & \b{99.1} /    72.7  & \b{99.8} /    76.4  &    90.4  /    52.3  & \b{99.8} /    91.0  &    96.4  /    90.3  &    80.3  /    74.8  &    93.8  /    82.9  &    95.3  /    78.6  \\
& \multicolumn{1}{c}{\ru"}   & inpainted     && \b{100.} / \b{99.4} & \b{100.} / \b{99.6} &    96.7  /    77.8  & \b{99.4} /    92.8  & \b{99.9} / \b{99.2} & \b{98.7} /    87.5  & \b{99.9} / \b{98.4} & \b{98.9} /    94.4  &    89.4  /    81.2  &    97.5  /    92.0  &    98.0  /    92.2  \\
& \multicolumn{1}{c}{\ru"}   & inpainted+    && \b{99.9} /    98.8  & \b{100.} / \b{99.7} & \b{99.7} / \b{95.9} & \b{99.6} /    96.8  & \b{100.} / \b{99.6} &    97.9  /    85.3  & \b{99.3} /    95.1  &    98.0  /    95.0  & \b{96.0} / \b{89.8} & \b{99.4} / \b{96.5} & \b{99.0} /    95.2  \\
& \multicolumn{1}{c}{\ru"}   & inpainted++   && \b{100.} / \b{99.6} & \b{100.} / \b{99.8} & \b{99.7} / \b{95.7} & \b{99.9} / \b{98.2} & \b{100.} / \b{99.7} & \b{99.3} / \b{92.3} & \b{99.9} / \b{98.9} & \b{99.0} /    95.6  & \b{95.8} /    86.3  & \b{99.7} / \b{97.4} & \b{99.3} / \b{96.4} \\
    \cmidrule(lr){1-3} \cmidrule(r){5-9} \cmidrule(r){10-11} \cmidrule(r){12-14} \cmidrule(lr){15-15}
    & Method & Training setting && Midjourney  &  SDXL   & DALL·E 2  &  DALL·E 3  &  Firefly  &  FLUX  & SD 3.5  & Facebook & Reddit & Twitter & NLL$\downarrow$/ECE$\downarrow$ \\ 
    \cmidrule(lr){1-3} \cmidrule(r){5-9} \cmidrule(r){10-11} \cmidrule(r){12-14} \cmidrule(lr){15-15}
\multirow{8}{*}{\rotatebox[origin=c]{90}{NLL / ECE}} 
& \ru  paired by text    & $-$           &&    1.96  /    .418  &    0.72  /    .218  &    3.60  /    .496  &    1.71  /    .416  &    2.95  /    .484  &    2.62  /    .473  &    1.42  /    .327  &    1.00  /    .273  &    1.77  /    .324  &    1.31  /    .310  &    1.91  /    .374  \\
& \ru  reconstructed     & $-$           && \b{0.00} / \b{.003} & \b{0.00} / \b{.001} &    3.33  /    .469  &    0.83  /    .240  &    1.56  /    .368  &    1.46  /    .354  &    0.38  /    .115  &    0.13  / \b{.025} &    1.20  /    .192  &    0.43  /    .082  &    0.93  /    .185  \\
& \ru  reconstructed     & inpainted     &&    0.01  /    .008  &    0.01  /    .008  &    1.16  /    .353  &    0.40  /    .197  &    0.51  /    .222  &    0.79  /    .290  &    0.23  /    .107  & \b{0.12} /    .032  &    0.73  /    .153  &    0.29  /    .066  &    0.42  /    .144  \\
& \ru  self-conditioned  & $-$           &&    0.08  /    .031  &    0.02  /    .008  &    1.48  /    .399  &    0.59  /    .234  &    0.27  /    .114  &    1.22  /    .379  &    0.04  /    .021  &    0.31  /    .084  &    0.92  /    .237  &    0.40  /    .078  &    0.53  /    .158  \\
\cdashline{2-3} \cdashline{5-15}
& \ru  self-conditioned  & cutmix/mixup  &&    0.14  /    .063  &    0.06  /    .028  &    1.66  /    .440  &    0.67  /    .268  &    0.48  /    .238  &    1.82  /    .452  &    0.22  /    .103  &    0.29  /    .076  &    0.77  /    .162  &    0.48  /    .141  &    0.66  /    .197  \\
 & \multicolumn{1}{c}{\ru"}   & inpainted     &&    0.02  /    .016  &    0.02  /    .017  &    0.49  /    .199  &    0.18  /    .085  &    0.04  /    .025  &    0.27  /    .117  &    0.05  /    .021  &    0.16  /    .070  &    0.39  /    .059  &    0.19  /    .033  &    0.18  /    .064  \\
& \multicolumn{1}{c}{\ru"}   & inpainted+    &&    0.04  /    .008  &    0.01  /    .006  &    0.12  / \b{.044} &    0.10  /    .037  & \b{0.02} / \b{.011} &    0.40  /    .149  &    0.14  /    .044  &    0.17  /    .032  & \b{0.25} / \b{.043} &    0.11  / \b{.028} &    0.14  /    .040  \\
& \multicolumn{1}{c}{\ru"}   & inpainted++   &&    0.02  /    .013  &    0.01  /    .011  & \b{0.10} /    .045  & \b{0.06} / \b{.031} &    0.02  /    .019  & \b{0.19} / \b{.084} & \b{0.04} / \b{.014} &    0.14  /    .039  &    0.28  /    .075  & \b{0.09} /    .047  & \b{0.10} / \b{.038} \\
    \bottomrule
    \end{tabular}
    }
  
    \vspace{-1.5mm}
    \caption{Ablation study. We compare several forms of content alignment and content augmentation. Performance are in terms of AUC/Accuracy (top) and ECE/NLL (bottom). Note that all variants share a standard augmentation (blurring + JPEG compression) as proposed in \cite{Wang2020cnn}. For content alignment we consider image pairing strategies described in Fig.~\ref{fig:training_approaches}: text-driven generation, reconstruction through autoencoder, and our proposal using self-conditioned images. For the last solution we test several forms of augmentation: a standard cutmix/mixup, and three proposed strategies based on inpainting described in Sec. \ref{sec:effect_aug}.
    Bold underlines the best performance for each column with a margin of 1\%.}
    \label{tab:results_abl_1}
\end{table*}

\section{Proposed Method}
\label{sec:proposed}

To realize and test our bias-free training paradigm we:
\begin{itemize}
\item build a dataset consisting of real and generated fake images, where the latter are well aligned with their real counterparts but include the forensic artifacts of the diffusion-based generation process. The dataset is created starting from the images collected from the training set of MS-COCO dataset \cite{Lin2014microsoft}, for a total of 51,517 real images. 
It is then enriched through several forms of augmentation, including locally inpainted images, and comprises eventually 309,102 generated images. 
\item use this aligned dataset to fine-tune end-to-end a Vision Transformer (ViT)-based model. 
Specifically, we adopt a variant of the ViT network proposed in \cite{Darcet2024vision} with four registers and use the pretraining based on the self-supervised learning method DINOv2 \cite{Oquab2024dinov2}.
During training, we avoid resizing the image and rely on large crops of $504\times504$ pixels. At inference time, we also extract crops of the same dimension from the image (if the image is larger we average the results of multiple crops).
\end{itemize}

\begin{figure}[t!]
\centering
{\footnotesize
\setlength{\tabcolsep}{1pt}
\begin{tabular}{cccc}
    \includegraphics[height=0.31\linewidth]{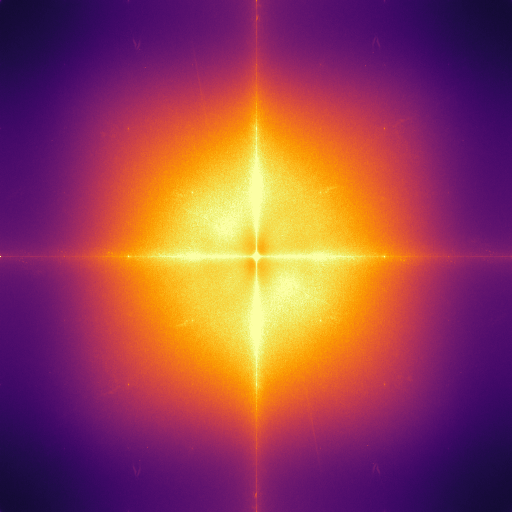}  &
    \includegraphics[height=0.31\linewidth]{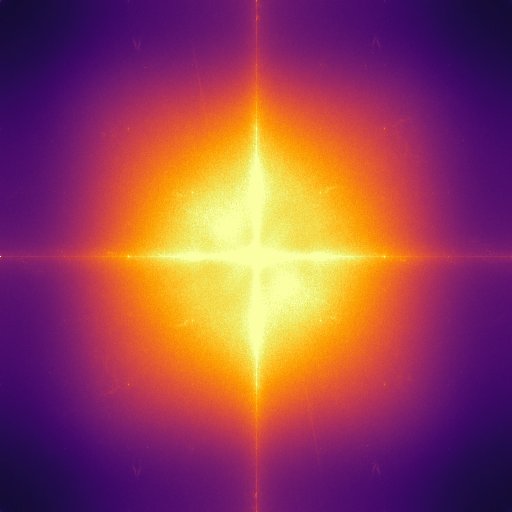} &
    \includegraphics[height=0.31\linewidth]{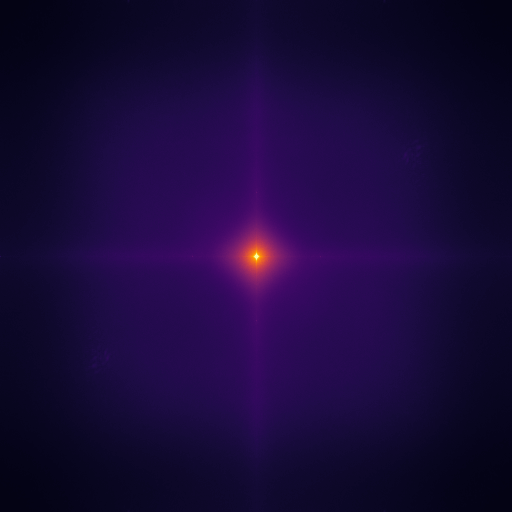} &
    ~\includegraphics[height=0.31\linewidth]{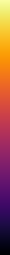} \\
    (a) & (b) & (c) &
\end{tabular}
}
    \caption{Power spectra computed by averaging (2000 images) the differences between: (a) real and reconstructed images, (b) real and self-conditioned images, and (c) reconstructed and self-conditioned images.
    We can observe that the self-conditioned generation embeds forensic artifacts even at lower frequencies compared to reconstructed images. This means that it is possible to better exploit such inconsistencies to distinguish real from fakes.}
    \label{fig:fft}
\end{figure}

\noindent
To ensure fake images semantically match the content of real images, we exploit the conditioning mechanism of Stable Diffusion models that allows us to control the synthesis process through a side input, which can be a class-label, a text or another image. The side input is firstly projected to an intermediate representation by a domain specific encoder, and then feeds the intermediate layers of the autoencoders for denoising in the embedding space. After several denoising steps, a decoder is used to obtain the conditioned synthetic image from embedded vector (See Fig.~\ref{fig:teaser}). In our self-conditioned generation, we use the inpainting diffusion model of Stable Diffusion 2.1 \cite{Rombach2022stable}, that has three side inputs: the reference image, a binary mask of the area to inpaint, and a textual description. Using an empty mask, we induce the diffusion steps to regenerate the input, that is, to generate a new image with exactly the same content as the input image. 
For the content augmentation process, we use the Stable Diffusion 2.1 inpainting method to replace an object with a new one, chosen from the same category or from a different one. Moreover, as shown in Fig.~\ref{fig:augmentation}, besides the default inpainting, which regenerates the whole image, we consider also a version where the original background is restored. Note that during training, we balance the real and fake class taking an equal number of images from each.

In the following, we present our ablation study. To avoid dataset bias we use WildRF and Synthbuster.
In addition, we test on 1000 FLUX and 1000 Stable Diffusion 3.5 images, which are some of the latest synthetic generators.

\subsection{Influence of content alignment} 
In Tab.~\ref{tab:results_abl_1} we show the performance achieved with different dataset alignment strategies, as described in Fig.~\ref{fig:training_approaches}.
Note that all variants are trained with standard augmentations, including blurring and JPEG compression, as proposed in \cite{Wang2020cnn}.
From the Table, we can observe that there is a large gain in terms of balanced accuracy ($\simeq$20\%) when moving from a text-driven generation (first row) to a solution where real and fake images share the semantic content, both reconstructed and self-conditioned.
The proposed solution that uses a diffusion pass demonstrates further improvement on average across all the evaluation metrics. This is highlighted in Fig.~\ref{fig:fft}, where we show the power spectra evaluated by averaging the difference between the real and reconstructed images and the real and self-conditioned images.
We observe that self-conditioned generation introduces forensic artifacts even at the lowest frequencies, indicating a detector trained on such images can exploit inconsistencies on a broader range of frequencies. 

\begin{figure}[t!]
    \centering
    \includegraphics[width=0.32\linewidth, page=1, trim=12 0 0 0,clip]{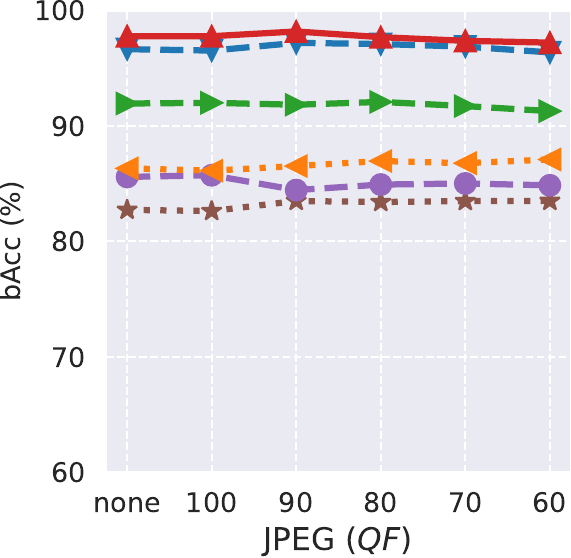}
    \includegraphics[width=0.324\linewidth, page=2, trim=12 0 0 0,clip]{figures/results_robustness.pdf}
    \includegraphics[width=0.32\linewidth, page=3, trim=12 0 0 0,clip]{figures/results_robustness.pdf}
    \\[1mm]
    \includegraphics[width=1.0\linewidth, page=4]{figures/results_robustness.pdf} 
    \caption{Robustness analysis in terms of balanced Accuracy carried out on nine generators of the Synthbuster dataset.}
    \label{fig:robustness}
\end{figure}

\subsection{Effect of content augmentation} 
\label{sec:effect_aug}
We also analyze the effect of different content augmentation strategies (Fig.~\ref{fig:augmentation}). We consider standard operations like cut-mix \cite{Yun2019cutmix} and mix-up \cite{Zhang2018mixup} and compare them with our proposed solutions that include three variants: 
\begin{itemize}
\item  {\em inpainted}, we replace an object with another from the same category plus the version where the background is substituted with pristine pixels (effectively a local image edit);
\item {\em inpainted+}, we replace an object with another from both the same and a different category plus the corresponding versions where the background is substituted with pristine pixels;
\item {\em inpainted++}, we further add some more standard augmentation operations, such as scaling, cut-out, noise addition, and jittering.
\end{itemize}
Overall, it is evident from Tab.~\ref{tab:results_abl_1} that augmentation plays a 
critical role in enhancing model generalization and this can be appreciated especially by looking at balanced accuracy and calibration measures. 
In fact, 
adding inpainting to reconstruction increases the accuracy from 80.7 to 83.9, while the joint use of self-conditioning and inpainting grants a significant extra gain, reaching 92.2 or even 96.4 with \textit{inpainted++}.
The most significant gains are observed on DALL·E 2, DALL·E 3 and FLUX that, probably, differ the most from Stable 2.1 in terms of architecture and hence require a stronger augmentation strategy to generalize. 

\begin{table}[t!]
    \resizebox{1.\linewidth}{!}{\small
    \begin{tabular}{lc C{10mm}C{10mm}C{10mm}C{10mm} }
    \toprule
    \b{Architecture} & \b{FT} &  \b{AUC}$\uparrow$ & \b{bAcc}$\uparrow$ & \b{NLL}$\downarrow$ & \b{ECE}$\downarrow$ \\
\cmidrule(lr){1-2} \cmidrule(r){3-6} 
\ru  DINOv2+reg       & LP    &    80.8  &    68.5  &    0.58  &    .141  \\
\ru  DINOv2+reg       & e2e   & \b{99.0} & \b{95.2} & \b{0.14} & \b{.040} \\
\ru  DINOv2           & e2e   & \b{98.4} &    91.1  &    0.24  &    .077  \\
\ru  SigLIP           & e2e   &    95.4  &    89.9  &    0.28  &    .066  \\
    \bottomrule
    \end{tabular}
    }
    \vspace{-1.5mm}
    \caption{We compare our solution, DINOv2+reg trained end-to-end, with linear probing (LP) and also consider alternative architectures, basic DINOv2 and SigLIP.}
    \label{tab:results_abl_arc}
\end{table}

\begin{table}[t!]
    \resizebox{1.\linewidth}{!}{\small
    \begin{tabular}{cl C{10mm}C{10mm}C{10mm}C{10mm} }
    \toprule
    \b{Architecture} &  \b{Training Set} & \b{AUC}$\uparrow$ & \b{bAcc}$\uparrow$ & \b{NLL}$\downarrow$ & \b{ECE}$\downarrow$ \\
\cmidrule(lr){1-2} \cmidrule(r){3-6}
\multirow{3}{*}{CLIP/ViT} 
 & \ru ProGAN \cite{Wang2020cnn}           &    54.7  &    45.2  &    4.85  &    .525  \\
 & \ru LDM \cite{Corvi2023detection}       &    63.9  &    48.5  &    3.39  &    .487  \\
 & \ru Ours                                & \b{75.2} & \b{73.9} & \b{0.66} & \b{.225} \\ \cmidrule(lr){1-2} \cmidrule(r){3-6}
\multirow{3}{*}{RINE }
 & \ru ProGAN \cite{Wang2020cnn}           &    66.1  &    65.0  &    5.43  &    .312  \\
 & \ru LDM \cite{Corvi2023detection}       &    83.0  &    75.7  &    0.53  &    .132  \\
 & \ru Ours                                & \b{89.9} & \b{83.2} & \b{0.39} & \b{.089} \\ 
    \bottomrule
    \end{tabular}
    }
    \vspace{-1.5mm}
    \caption{Ablation study on the influence of the training data (ProGAN, Latent Diffusion and our dataset) on methods used for AI-generated image detection: CLIP \cite{Radford2021learning} and RINE \cite{Koutlis2024leveraging}.}
    \label{tab:results_abl_2}
\end{table}

\begin{table*}[t!]
{\small
\centering
\resizebox{1.0\linewidth}{!}{
    \begin{tabular}{l  C{14mm}C{14mm}C{14mm}C{14mm}C{14mm} C{14mm}C{14mm} C{14mm}C{14mm}C{14mm} C{21mm}}
    \toprule
   \multirow{2}{*}{\b{bAcc(\%)}$\uparrow$/\b{NLL}$\downarrow$} & \multicolumn{5}{c}{\b{Synthbuster}} & \multicolumn{2}{c}{\b{New Generators}} & \multicolumn{3}{c}{\b{WildRF}} & \b{AVG} \\
     \cmidrule(r){2-6} \cmidrule(r){7-8} \cmidrule(lr){9-11} \cmidrule(lr){12-12}
     &  Midjourney  &  SDXL   & DALL·E 2  &  DALL·E 3  &  Firefly  &  FLUX & SD 3.5 & Facebook & Reddit & Twitter &  bAcc$\uparrow$/NLL$\downarrow$ \\  
     \cmidrule(lr){1-1} \cmidrule(r){2-6} \cmidrule(r){7-8} \cmidrule(lr){9-11} \cmidrule(lr){12-12}
\ru CNNDetect         &    49.5  /    8.45  &    49.8  /    6.90  &    50.2  /    5.75  &    49.5  /    12.9  &    50.3  /    3.66  &    49.5  /    10.1  &    50.0  /    5.30  &    50.0  /    9.84  &    50.7  /    6.66  &    50.1  /    8.64  &    50.0  /    7.83  \\
\ru DMID              & \b{100.} / \b{0.00} & \b{99.7} / \b{0.01} &    50.1  /    5.99  &    50.0  /    7.08  &    51.0  /    1.72  &    63.7  /    1.27  & \b{99.9} / \b{0.01} &    87.8  /    0.52  &    74.3  /    1.82  &    79.1  /    0.85  &    75.6  /    1.93  \\
\ru LGrad             &    57.7  /    6.88  &    58.5  /    6.81  &    55.6  /    7.10  &    47.9  /    7.58  &    47.4  /    7.50  &    54.9  /    7.11  &    51.9  /    7.24  &    66.6  /    3.74  &    57.8  /    4.72  &    45.7  /    5.31  &    54.4  /    6.40  \\
\ru UnivFD            &    52.4  /    2.35  &    68.0  /    1.15  &    83.5  /    0.43 &    47.3  /    3.94  &    90.7  /    0.24  &    48.4  /    3.45  &    69.3  /    1.07  &    48.8  /    3.06  &    59.5  /    1.37  &    56.0  /    2.01  &    62.4  /    1.91  \\
\ru DeFake            &    69.7  /    0.72  &    76.3  /    0.56  &    64.0  /    0.92  &    84.9  /    0.36  &    72.4  /    0.63  &    79.2  /    0.46  &    81.2  /    0.42  &    66.3  /    0.89  &    65.9  /    0.82  &    63.4  /    0.94  &    72.3  /    0.67  \\
\ru DIRE              &    49.7  /    15.3  &    49.9  /    15.3  &    50.0  /    15.3  &    50.0  /    15.3  &    49.9  /    15.3  &    50.0  /    15.3  &    50.0  /    15.3  &    51.9  /    4.98  &    79.5  /    2.15  &    56.7  /    4.39  &    53.7  /    11.9  \\
\ru AntifakePrompt    &    70.4  / ~~~~-~~~ &    84.7  / ~~~~-~~~ &    65.5  / ~~~~-~~~ &    86.0  / ~~~~-~~~ &    70.0  / ~~~~-~~~ &    59.6  / ~~~~-~~~ &    60.7  / ~~~~-~~~ &    69.7  / ~~~~-~~~ &    68.9  / ~~~~-~~~ &    78.0  / ~~~~-~~~ &    71.3  / ~~~~-~~~ \\
\ru NPR               &    44.9  /    16.6  &    50.3  /    16.2  &    50.2  /    16.2  &    ~0.6  /    29.9  &    ~0.4  /    47.3  &    50.3  /    16.2  &    50.3  /    16.2  &    50.0  /    32.2  &    78.3  /    9.39  &    51.8  /    25.2  &    42.7  /    22.5  \\
\ru FatFormer         &    44.4  /    5.22  &    66.7  /    2.76  &    54.1  /    3.64  &    35.9  /    6.90  &    60.1  /    3.59  &    39.4  /    6.10  &    49.1  /    5.06  &    54.7  /    4.54  &    69.5  /    2.54  &    54.8  /    4.40  &    52.9  /    4.48  \\
\ru FasterThanLies    &    61.3  /    2.98  &    71.1  /    1.79  &    50.8  /    5.15  &    53.5  /    3.79  &    55.2  /    4.40  &    53.8  /    4.10  &    53.7  /    3.76  &    46.2  /    3.32  &    51.0  /    3.99  &    53.9  /    3.31  &    55.1  /    3.66  \\
\ru RINE              &    54.6  /    5.03  &    71.8  /    1.99  &    82.2  /    0.77  &    45.3  /    20.5  &    91.2  /    0.36  &    46.7  /    10.1  &    81.3  /    1.22  &    52.8  /    6.51  &    67.7  /    2.46  &    56.0  /    5.23  &    65.0  /    5.43  \\
\ru AIDE              &    57.5  /    0.95  &    68.4  /    0.70  &    34.9  /    1.34  &    33.7  /    1.38  &    24.8  /    2.00  &    62.9  /    0.82  &    63.3  /    0.82  &    56.9  /    0.94  &    72.1  /    0.62  &    57.3  /    1.01  &    53.2  /    1.06  \\
\ru LaDeDa            &    50.7  /    24.8  &    50.7  /    24.8  &    50.5  /    24.8  &    41.1  /    25.4  &    47.4  /    25.6  &    50.5  /    24.8  &    50.7  /    24.8  &    70.3  /    7.19  &    74.7  /    7.93  &    59.6  /    9.40  &    54.6  /    19.9  \\
\ru C2P-CLIP          &    52.8  /    1.10  &    77.7  /    0.48  &    55.6  /    0.99  &    63.2  /    0.73  &    59.5  /    0.89  &    50.1  /    1.30  &    60.9  /    0.93  &    54.4  /    0.97  &    68.4  /    0.67  &    57.4  /    0.91  &    60.0  /    0.90  \\
\ru CoDE     &    76.9  /    0.82  &    75.2  /    0.81  &    54.6  /    2.44  &    73.2  /    0.98  &    58.6  /    2.00  &    59.8  /    1.97  &    67.7  /    1.27  &    70.0  /    0.97  &    66.1  /    1.29  &    70.9  /    1.01  &    67.3  /    1.36  \\
\cmidrule(lr){1-1} \cmidrule(r){2-6} \cmidrule(r){7-8} \cmidrule(lr){9-11} \cmidrule(lr){12-12}
\ru B-Free (ours)     & \b{99.6} /    0.02  & \b{99.8} / \b{0.01} & \b{95.6} / \b{0.10} & \b{98.2} / \b{0.06} & \b{99.7} / \b{0.02} & \b{92.3} / \b{0.19} &    98.9  /    0.04  & \b{95.6} / \b{0.14} & \b{86.2} / \b{0.28} & \b{97.3} / \b{0.09} & \b{96.3} / \b{0.10} \\
 \bottomrule
\end{tabular}
}
\vspace{-1.5mm}
\caption{Comparison with SoTA methods in terms of balanced Accuracy and balanced NLL across different generators. Note that AntifakePrompt \cite{Chang2023antifake} provides only hard binary labels hence calibration measures cannot be computed. Bold underlines the best performance for each column with a margin of 1\%.}
\label{tab:RES_ACC_1}
}
\end{table*}

In Fig.~\ref{fig:robustness}, we analyze the impact of our content augmentation, assessing robustness under various operations: JPEG compression, resizing, and blurring. 
All three proposed variants of augmentation offer a clear advantage, especially when resizing is applied. The joint use of self-conditioning and inpainting results in the most robust approach.

\subsection{Influence of training data and architecture}
We conduct additional experiments to gain deeper insights into the impact of the chosen architecture and the proposed training data on the same datasets shown in Tab.~\ref{tab:results_abl_1}. 

First we compare our adopted model, DINOv2+reg trained end-to-end, with an alternative fine-tuning strategy that involves training only the final linear layer, known as linear probing (LP) that is largely adopted in the literature \cite{Ojha2023towards, Cozzolino2024raising}. From Tab.~\ref{tab:results_abl_arc} we can see that this latter solution does not perform well. One possible explanation is that features from last layer capture high-level semantics, while our dataset is built to exploit low-level artifacts that derive from first and intermediate layers \cite{Corvi2023detection, Koutlis2024leveraging}. In the same Table we compare DINOv2+reg with the basic DINOv2 \cite{Oquab2024dinov2} architecture and SigLIP \cite{Zhai2023sigmoid} and we can observe that DINOv2 with the use of registers achieves the best performance, probably thanks to the fact that it avoids to discard local patch information \cite{Darcet2024vision}.

Then, in Tab.~\ref{tab:results_abl_2}, we consider a CLIP-based model and the architecture of RINE, and vary the training dataset by including two well known datasets largely used in the literature, one based on ProGAN \cite{Wang2020cnn} and the other on Latent Diffusion \cite{Corvi2023detection}. 
We note that our training paradigm achieves the best performance over all the metrics with a very large gain (RINE increases the accuracy from 65\% to 83.2\%).

\begin{figure*}[t!]
    \includegraphics[width=1.0\linewidth,clip,trim=0 0 0 0]{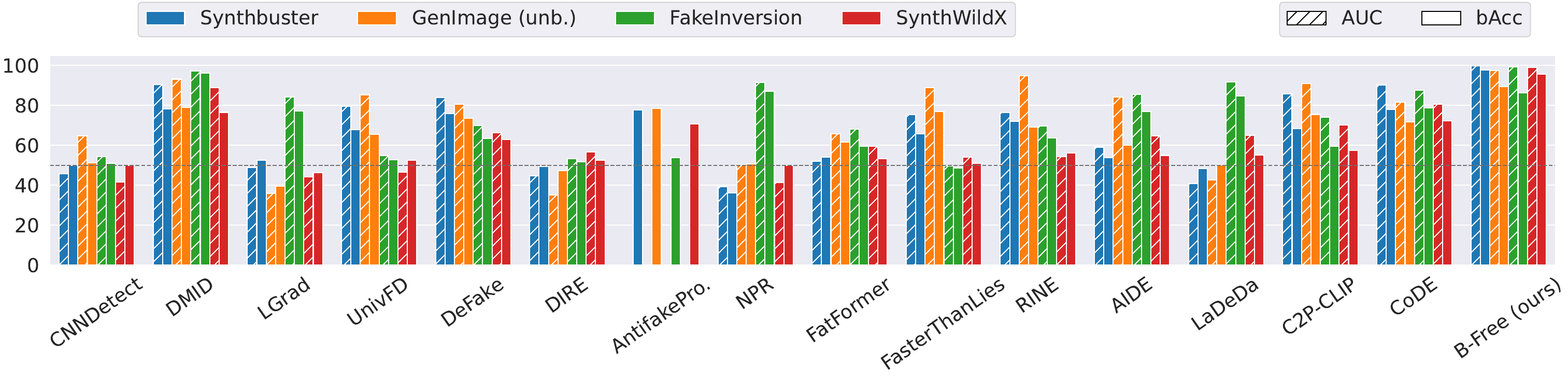}
    \vspace{-1.7\baselineskip}
    \caption{Average performance in term of AUC and bAcc on four datasets: Synthbuster, GenImage, FakeInversion, SynthWildX.}
    \label{fig:enter-label}
\end{figure*}

\section{Comparison with the State-of-The-Art}
\label{sec:comparison}

In this Section, we conduct a comparison with SoTA methods on 27 diverse synthetic generation models. To ensure fairness, we include only SoTA methods with publicly available code and/or pre-trained models. The selected methods are listed in Table~\ref{tab:methods} and are further described in the supplementary material together with additional experiments. 
For all the experiments now on, \textit{ours} refers to the detector trained using \textit{inpainted++} augmentation.

\begin{table}[t!]
    \centering
    \resizebox{1.0\linewidth}{!}{\small
    \begin{tabular}{clp{3.1cm}cc}
    \toprule
        \textbf{Ref.} & \textbf{Acronym}& \textbf{Training Real/Fake} & \textbf{Size (K)} & \textbf{Aug.} \\
    \midrule
    \cite{Wang2020cnn}            & CNNDetect       & LSUN / ProGAN & 360 / 360   & \checkmark   \\
    \cite{Corvi2023detection}     & DMID            & COCO, LSUN / Latent & 180 / 180 & \checkmark  \\
    \cite{Tan2023learning}        & LGrad           & LSUN / ProGAN & 72 / 72     & \checkmark  \\
    \cite{Ojha2023towards}        & UnivFD          & LSUN / ProGAN & 360 / 360   & \checkmark \\
    \cite{Sha2023defake}          & DeFake          & COCO / SD & 20 / 20 &  \\
    \cite{Wang2023dire}           & DIRE            & LSUN-Bed / ADM & 40 / 40 &  \\
    \cite{Chang2023antifake}      & AntifakePrompt  &  COCO / SD3,SD2-inp & 90 / 60 & \checkmark \\
    \cite{Tan2024rethinking}      & NPR             & LSUN / ProGAN & 72 / 72     &       \\
    \cite{Liu2024forgery}         & FatFormer       &  LSUN / ProGAN & 72 / 72  &    \\
    \cite{Lanzino2024faster}      & FasterThanLies  & COCO / SD & 108 / 542       & \checkmark  \\
    \cite{Koutlis2024leveraging}  & RINE            & LSUN / ProGAN & 72 / 72   & \checkmark  \\
    \cite{Yan2024sanity}          & AIDE            & ImageNet / SD 1.4  & 160 / 160 & \checkmark  \\
    \cite{Cavia2024real}          & LaDeDa          & LSUN / ProGAN & 360 / 360 &   \\ 
    \cite{Tan2024c2p}             & C2P-CLIP        & LSUN / ProGAN & 72 / 72 & \checkmark  \\ 
    \cite{Baraldi2024contrastive} & CoDE            & \multicolumn{1}{R{3.1cm}}{LAION / SD1.4,~SD2.1, SDXL, DeepF.~IF} & 2.3M / 9.2M & \checkmark\\
    \cdashline{1-5}
    \rule{0pt}{10pt}
    & B-Free (ours) & COCO / SD2.1 & 51 / 309 & \checkmark\\
    \bottomrule
    \end{tabular}
    }
    \vspace{-1.5mm}
    \caption{AI-generated image detection methods used for comparison and whose code is publicly available. We specify source and size of the training dataset, and whether augmentation is applied.}
    \label{tab:methods}
\end{table}
A first experiment is summarized in Tab.~\ref{tab:RES_ACC_1} with results given in terms of  balanced accuracy and NLL. Most of the methods struggle to achieve a good accuracy, especially on more recent generators. Instead, \NAME~obtains a uniformly good  performance on all generators, irrespective of the image origin, whether they are saved in raw format or downloaded from social networks, outperforming the second best (see last column) by $+20.7\%$ in terms of  bAcc. 
Then, we evaluate again all methods on GenImage (unbiased), FakeInversion \cite{Cazenavette2024fakeinversion}, and SynthWildX \cite{Cozzolino2024raising}. As these datasets encompass multiple generators, we only report the average performance in Tab.~\ref{tab:RES_ACC_2}. On these additional datasets, most methods provide unsatisfactory results, especially in the most challenging scenario represented by SynthWildX, with images that are shared over the web. The proposed method performs well on all datasets, just a bit worse on FakeInversion.  
Finally in Fig.~\ref{fig:enter-label} we study how AUC compares with balanced accuracy for all the methods over several datasets. We observe that some methods, like NPR and LGrad, present a clear non-uniform behavior, with very good performance on a single dataset and much worse on the others. This seems to suggest that these methods may not be truly detecting forensic artifacts, instead are rather exploiting intrinsic biases within the dataset. Differently, the proposed method presents a uniform performance across all datasets and a small loss between AUC and accuracy.

\begin{figure}[t!]
    \begin{subfigure}[t]{0.33\textwidth}
        \includegraphics[width=1.0\textwidth,clip,trim=0 0 0 0]{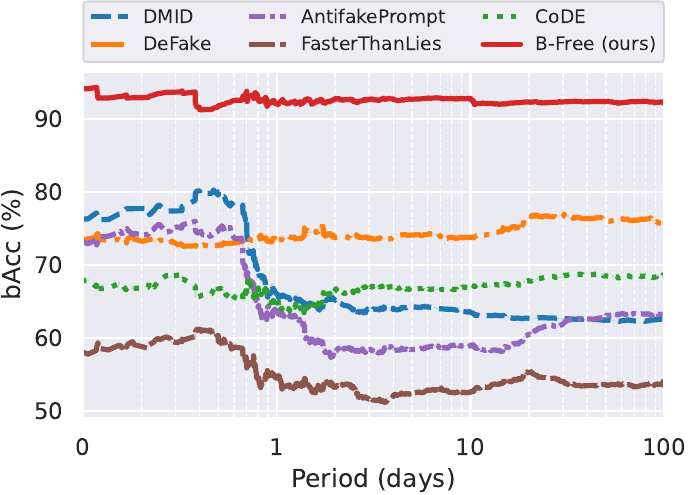}
    \end{subfigure}
    \hfill
    \begin{subfigure}[t]{0.12\textwidth}
        \includegraphics[width=1.0\textwidth,page=4,clip,trim=0 95 740 0]{figures/figures.pdf}
    \end{subfigure}
    \caption{
    Results of SoTA detectors on real and fake images that went viral on internet, analyzing multiple web-scraped versions of each image. The performance is in terms of balanced accuracy evaluated from the initial online post (Log scale). Only detectors with an accuracy over 50\% are shown. }
    \label{fig:time}
\end{figure}

\bb{Analysis on content shared on-line.} 
Distinguishing real from synthetic images on social networks may be especially challenging due to the presence of multiple re-posting that impair image quality over time. A recent study conducted in \cite{Karageogiou2024evolution} analyzed the detector behavior on different instances of an image shared online, showing that the performance degrades noticeably in time due to repeated re-posting. To better understand the impact of our augmentation strategies on such images, we collected a total of 1400 real/fake images that went viral on the web, including several versions of the same real or fake image.

Fig.~\ref{fig:time} illustrates the accuracy, which is evaluated over a 100-day period from the time of initial publication, with times on a logarithmic scale. We compare our proposal with the best performing SoTA methods.
We can notice that the performance drops after only one day, after which most competitors are stuck below 70\%, with the exception of DeFake that achieves around 75\%. Only the proposed method, which comprises more aggressive augmentation, is able to ensure an average accuracy around 92\% even after many days from the first on-line post.

\begin{table}[t!]
\resizebox{1.0\linewidth}{!}{\small
    \begin{tabular}{l  C{14mm}C{14mm}C{14mm} C{14mm}}
    \toprule
  \b{bAcc(\%)}$\uparrow$/\b{NLL}$\downarrow$ & \b{GenImage} & \b{FakeInver}. &  \b{SynthWildX} & \b{AVG} \\
   \cmidrule(lr){1-1}  \cmidrule{2-4} \cmidrule(lr){5-5}
\ru CNNDetect         &    51.3  /    7.88  &    50.9  /    7.94  &    50.0  /    8.08  &    50.7  /    7.96  \\
\ru DMID              &    79.0  /    1.66  & \b{96.1} / \b{0.25} &    76.6  /    0.82  &    83.9  /    0.91  \\
\ru LGrad             &    39.6  /    7.12  &    77.2  /    2.27  &    46.3  /    5.53  &    54.3  /    4.97  \\
\ru UnivFD            &    65.5  /    1.31  &    52.8  /    2.19  &    52.5  /    2.55  &    56.9  /    2.02  \\
\ru DeFake            &    73.7  /    0.74  &    63.3  /    0.95  &    62.9  /    0.98  &    66.6  /    0.89  \\
\ru DIRE              &    47.3  /    6.54  &    51.8  /    13.4  &    52.5  /    4.60  &    50.5  /    8.19  \\
\ru AntifakePrompt    &    78.5  / ~~~~-~~~ &    53.9  / ~~~~-~~~ &    70.8  / ~~~~-~~~ &    67.8  / ~~~~-~~~ \\
\ru NPR               &    50.7  /    25.3  &    87.0  /    4.96  &    49.9  /    28.2  &    62.6  /    19.5  \\
\ru FatFormer         &    61.5  /    3.99  &    59.7  /    3.45  &    53.3  /    4.75  &    58.2  /    4.06  \\
\ru FasterThanLies    &    77.0  /    1.23  &    48.6  /    3.64  &    50.9  /    3.40  &    58.8  /    2.76  \\
\ru RINE              &    69.1  /    2.57  &    63.6  /    4.84  &    56.2  /    6.07  &    63.0  /    4.49  \\
\ru AIDE              &    60.2  /    1.01  &    76.9  /    0.54  &    55.0  /    1.05  &    64.0  /    0.86  \\
\ru LaDeDa            &    50.2  /    29.2  &    84.7  /    3.03  &    55.1  /    10.2  &    63.3  /    14.1  \\
\ru C2P-CLIP          &    75.5  /    0.57  &    59.6  /    0.82  &    57.4  /    0.91  &    64.2  /    0.76  \\
\ru CoDE              &    71.7  /    1.43  &    78.8  /    0.74  &    72.3  /    0.95  &    74.2  /    1.04  \\
\cmidrule(lr){1-1}  \cmidrule{2-4} \cmidrule(lr){5-5}
\ru B-Free (ours)     & \b{89.3} / \b{0.27} &    86.2  /    0.32  & \b{95.6} / \b{0.14} & \b{90.4} / \b{0.24} \\
 \bottomrule
\end{tabular}
}
\vspace{-1.5mm}
\caption{Comparison with SoTA methods in terms of average performance in terms of balanced accuracy and NLL for three additional datasets: GenImage, FakeInversion and SynthWildX.}
\label{tab:RES_ACC_2}
\end{table}

\section{Limitations}
\label{sec:limitations}

The method proposed in this work is trained using fake images that are self-conditioned reconstructions from Stable Diffusion 2.1 model. If new generators will be deployed in the future that have a completely different synthesis process, then is it very likely that this approach will fail (the principles and ideas shared in this work may still hold). Further, being a data-driven approach it can be adversarially attacked by a malicious user. This is a very relevant issue that we plan to address in our future work.

\section{Conclusions}
\label{sec:conclusions}

In this paper, we propose a new training paradigm for AI-generated image detection. First of all, we empirically demonstrate the importance of pairing real and fake images by constraining them to have the same semantic content. This helps to better extract common artifacts shared across diverse synthetic generators.
Then we find that using aggressive data augmentation, in the form of partial manipulations, further boosts performance both in term of accuracy and of calibration metrics. This is extremely relevant especially when working in realistic scenarios, such as image sharing over social networks.
Our findings emphasize that careful dataset curation and proper training strategy can be more impactful compared to developing more complex algorithms. We hope this work will inspire other researchers in the forensic community to pursue a similar direction, fostering advancements in bias-free training strategies.

\paragraph{Acknowledgments.}
We gratefully acknowledge the support of this research by a Google Gift. In addition, this work has received funding from the European Union under the Horizon Europe vera.ai project, Grant Agreement number 101070093, and was partially supported by SERICS (PE00000014) under the MUR National Recovery and Resilience Plan, funded by the European Union - NextGenerationEU. 

\begin{appendix}
\section*{Supplementary Material}
\def\scalefactortab{0.76}

In this supplementary material, 
we report more details about our implementation (Sec.~\ref{implementation}).
Moreover, we briefly describe the state of the art methods we compare to (Sec.~\ref{methods}),
and give more details about the calibration metrics used in the experiments (Sec.~\ref{metrics}). 
We also provide additional ablation results in Sec.~\ref{ablation} and carry out further experiments on generalization (Sec.~\ref{generalization}) on additional publicly available datasets.
Furthermore, we provide more results on the robustness of our approach compared with SoTA methods (Sec.~\ref{robustness}).

\section{Implementation Details}
\label{implementation}

\noindent
\bb{Training strategy.}
The proposed model leverages the DINOv2+reg \cite{Oquab2024dinov2, Darcet2024vision} $504\times 504$ image embedding network as its backbone,
followed by a fully connected layer. The model is trained end-to-end using the binary cross-entropy loss function on an NVIDIA A100 GPU. The training process employs the ADAM optimizer with a learning rate of 1e-6, a weight decay of 1e-6, and a batch size of 24. During training, the balanced accuracy is evaluated on a validation set every 3435 iterations. Early stopping is applied to prevent overfitting: training is completed if the validation balanced accuracy does not improve by at least 0.1\% over five consecutive evaluations.

\vspace{2mm}
\noindent
\bb{Test strategy.} If the test image is less than 504 pixels, padding is applied after patch embedding.
Otherwise, we average the logit score over multiple crops to analyze the whole image.

\begin{figure}[t!]
    \centering
    \includegraphics[width=0.98\linewidth, page=5,clip, trim=170 50 160 0]{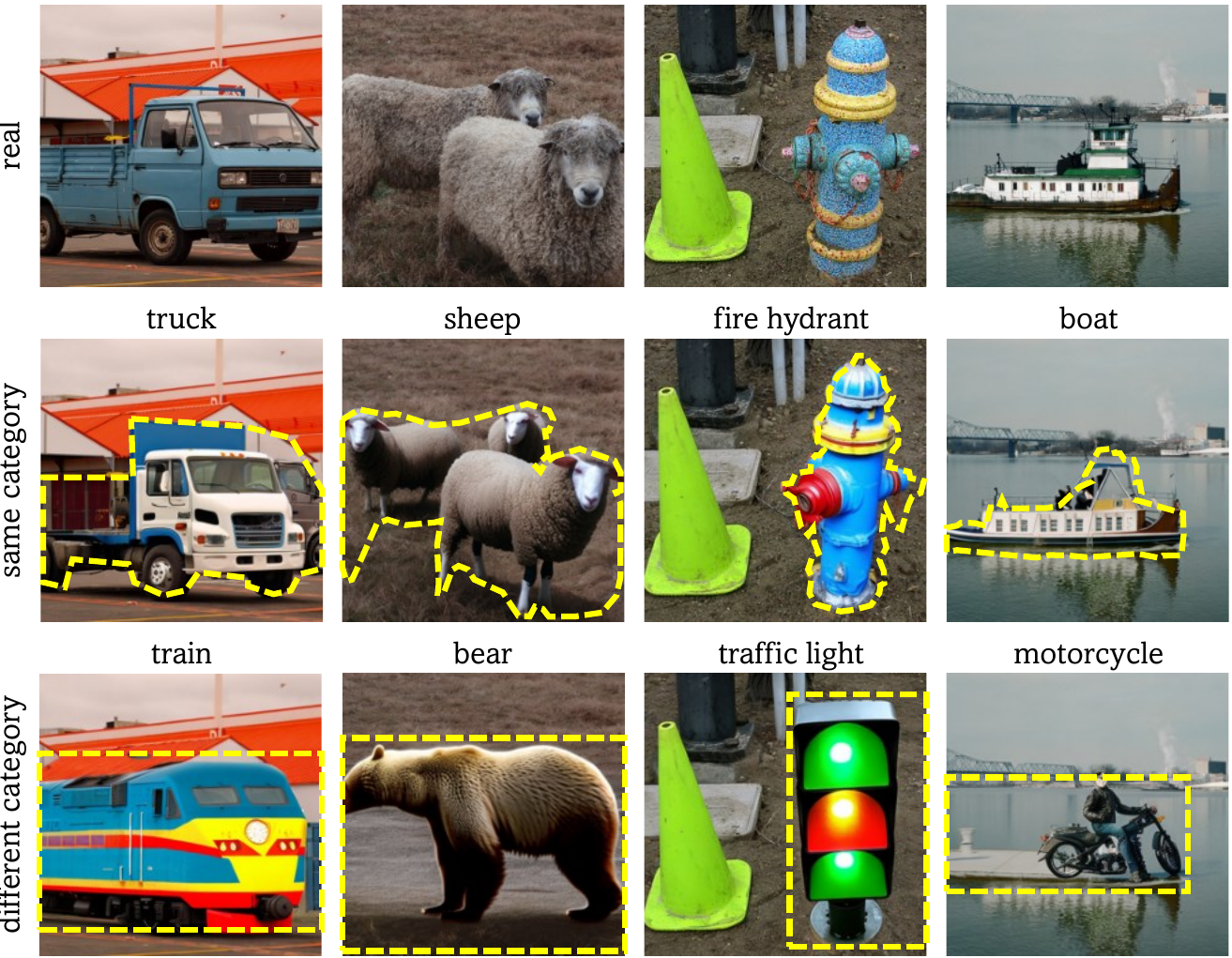}
 
     \caption{Examples of content augmented images from our training dataset. From real images (first row), we generate inpainted versions with the same content (second row) and different content (third row).}
    \label{fig:training_dataset}
\end{figure}

\vspace{2mm}
\noindent
\bb{Training Dataset.}
Here we give more deatils on how we built our dataset.
Starting from the MS-COCO training set, consisting of 118K images with 80 categories of objects, we first discarded images with licenses different than Creative Commons. 
Before editing the images, we extracted the largest central crop, which allows us to retain most of the semantic content of the original image.
We discarded images where objects are not present and ended up with a pristine source of 51,517 images.
For content augmentation, we replaced the selected object with an object generated from the same category using the COCO segmentation mask, and from a different category using a rectangular box.
We took care to not affect too much the realism of the content, so for the ``different category'' case the object is changed with one from a similar category, that belongs to the same COCO supercategory. In this scenario, the only exception is the category \textit{person}, which does not have a supercategory and it is therefore replaced with a random object. 
As mentioned in the main paper, besides the default inpainting, we also consider a version where we take the pixels of the object from the generated image, and the pixels of the background from the original one. We did the same with the self-conditioned image, restoring the background with original pixels. Therefore, we ended up with six fake versions for each real image (Fig.~\ref{fig:augmentation}).

\section{SoTA Methods}
\label{methods}
Below we provide a brief description of the methods we included in the comparison in Section \ref{sec:comparison}. The training datasets used by these methods are indicated in Table \ref{tab:methods}.

\vspace{2mm}
\noindent
\bb{CNNDetect \cite{Wang2020cnn}.} This is a CNN-based detector built on ResNet50 (pre-trained on ImageNet) that adpots augmentation in the form of post-processing operations, such as blurring and compression.

\vspace{2mm}
\noindent
\bb{DMID \cite{Corvi2023detection}.} This work also relies on a ResNet-50, but it prevents down-sampling at the first layer so as to preserve the invisible
forensics clues as much as possible, and uses a stronger augmentation to increase robustness.

\begin{table*}[t!]
    \resizebox{1.\linewidth}{!}{\small
    \begin{tabular}{l C{14mm}C{14mm}C{14mm}C{14mm}C{14mm} C{14mm}C{14mm} C{14mm}C{14mm}C{14mm} C{22mm}}
    \toprule
     & \multicolumn{5}{c}{\b{Synthbuster}} & \multicolumn{2}{c}{\b{New Generators}} & \multicolumn{3}{c}{\b{WildRF}} & \b{AVG} \\
    \cmidrule(lr){1-1} \cmidrule(r){2-6} \cmidrule(r){7-8} \cmidrule(r){9-11} \cmidrule(lr){12-12}
Training dataset & Midjourney  &  SDXL   & DALL·E 2  &  DALL·E 3  &  Firefly  &  FLUX & SD 3.5 & Facebook & Reddit & Twitter & AUC$\uparrow$/bAcc$\uparrow$  \\ 
    \cmidrule(lr){1-1} \cmidrule(r){2-6} \cmidrule(r){7-8} \cmidrule(r){9-11} \cmidrule(lr){12-12}
\ru D$^3$ \cite{Baraldi2024contrastive}  & \b{99.8} /    85.3  & \b{100.} / \b{99.4} & \b{100.} /    87.5  & \b{100.} /    90.1  & \b{100.} /    89.7  & \b{98.8} /    63.1  & \b{99.9} / \b{96.4} & \b{98.3} /    88.8  & \b{95.4} /    86.7  &    98.1  /    88.2  & \b{99.0} /    87.5  \\
\ru Ours                                 & \b{99.9} / \b{98.8} & \b{100.} / \b{99.7} & \b{99.7} / \b{95.9} & \b{99.6} / \b{96.8} & \b{100.} / \b{99.6} & \b{97.9} / \b{85.3} & \b{99.3} /    95.1  & \b{98.0} / \b{95.0} & \b{96.0} / \b{89.8} & \b{99.4} / \b{96.5} & \b{99.0} / \b{95.2} \\
  \cmidrule(lr){1-1} \cmidrule(r){2-6} \cmidrule(r){7-8} \cmidrule(r){9-11} \cmidrule(lr){12-12}
Training dataset & Midjourney  &  SDXL   & DALL·E 2  &  DALL·E 3  &  Firefly  &  FLUX  & SD 3.5  & Facebook & Reddit & Twitter & NLL$\downarrow$/ECE$\downarrow$ \\ 
  \cmidrule(lr){1-1} \cmidrule(r){2-6} \cmidrule(r){7-8} \cmidrule(r){9-11} \cmidrule(lr){12-12}
\ru D$^3$ \cite{Baraldi2024contrastive}  &    0.33  /    .156  &    0.03  /    .016  &    0.27  /    .146  &    0.22  /    .120  &    0.22  /    .138  &    1.07  /    .350  & \b{0.11} /    .048  &    0.26  /    .101  &    0.32  /    .079  &    0.28  /    .105  &    0.31  /    .126  \\
\ru Ours                                 & \b{0.04} / \b{.008} & \b{0.01} / \b{.006} & \b{0.12} / \b{.044} & \b{0.10} / \b{.037} & \b{0.02} / \b{.011} & \b{0.40} / \b{.149} &    0.14  / \b{.044} & \b{0.17} / \b{.032} & \b{0.25} / \b{.043} & \b{0.11} / \b{.028} & \b{0.14} / \b{.040} \\
    \bottomrule
    \end{tabular}
    }
  
    \vspace{-1.5mm}
    \caption{Ablation study on the influence of the training data. We compare the proposal with the same architecture DINOv2+reg trained on a publicly available dataset, D$^3$ \cite{Baraldi2024contrastive}, that includes 4 generators from the Stable Diffusion family. Performance are presented in terms of AUC/Accuracy (top) and ECE/NLL (bottom).}
    \label{tab:results_abl_3}
\end{table*}

\vspace{2mm}
\noindent
\bb{LGrad \cite{Tan2023learning}.} This work is also based on a ResNet-50 classifier, but this is fed by a generalized artifacts representation of the image in the form of gradients. This representation is designed to more effectively capture the artifacts introduced by synthetic generators.

\vspace{2mm}
\noindent
\bb{UnivFD \cite{Ojha2023towards}.} It exploits pre-trained CLIP features through linear probing. Fine-tuning is carried out on the same dataset of real and GAN-generated images as in \cite{Wang2020cnn}.

\vspace{2mm}
\noindent
\bb{DeFake \cite{Sha2023defake}.}
Both images and their corresponding prompts are used and fed into the visual and textual encoders of CLIP. The extracted features are the input of a multilayer perceptron trained for binary detection.

\vspace{2mm}
\noindent
\bb{DIRE \cite{Wang2023dire}.} It uses the reconstruction error of a generative model as the input of a ResNet-50. In fact, this error is expected to be lower for synthetic images than for real ones.

\vspace{2mm}
\noindent
\bb{AntifakePrompt \cite{Chang2023antifake}.} It relies on a visual question-answering (VQA) tool, InstructBLIP. The VQA is used with a fixed question, ``Is this photo real?", and fine-tuned to provide accurate responses (``Yes" or ``No") using a soft prompt tuning technique. Note that the method provides hard binary predictions hence only accuracy can be computed. 

\vspace{2mm}
\noindent
\bb{NPR \cite{Tan2024rethinking}.} In this case a ResNet-50 is fed using 
a residual image computed as the difference between the original image and its interpolated
version. The idea is to exploit the artifacts related to the up-sampling process which is common in several generative models.

\begin{figure*}[t!]
    \includegraphics[width=1.0\linewidth,clip,trim=0 0 0 0]{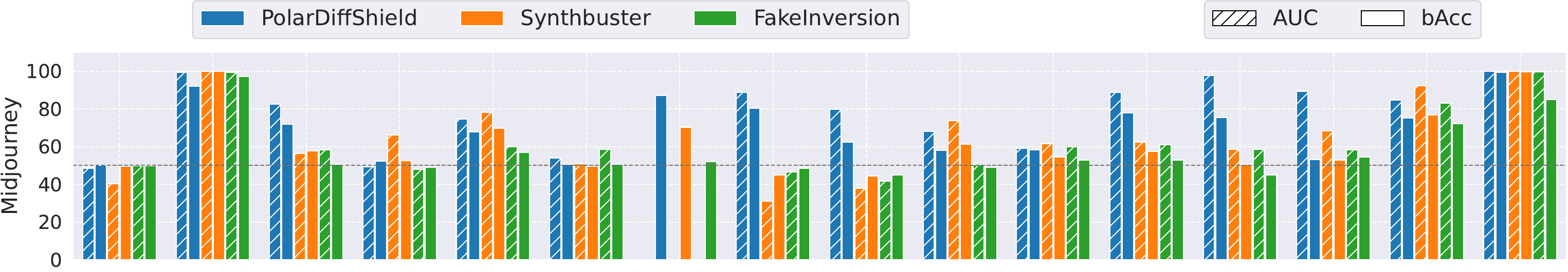} \\
    \includegraphics[width=1.0\linewidth,clip,trim=0 0 0 0]{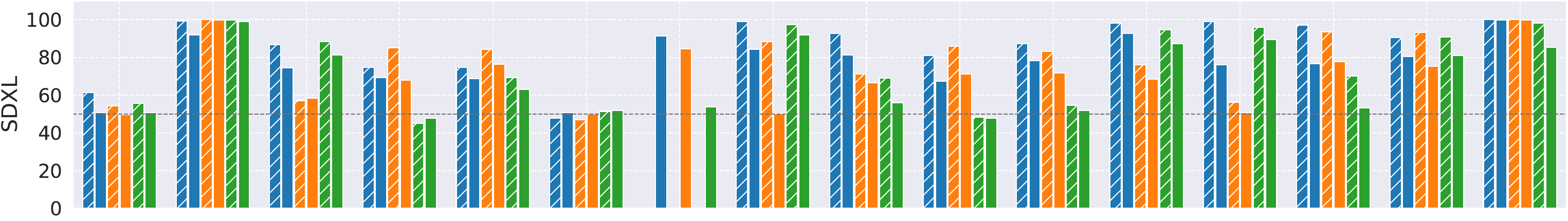}  \\
    \includegraphics[width=1.0\linewidth,clip,trim=0 0 0 0]{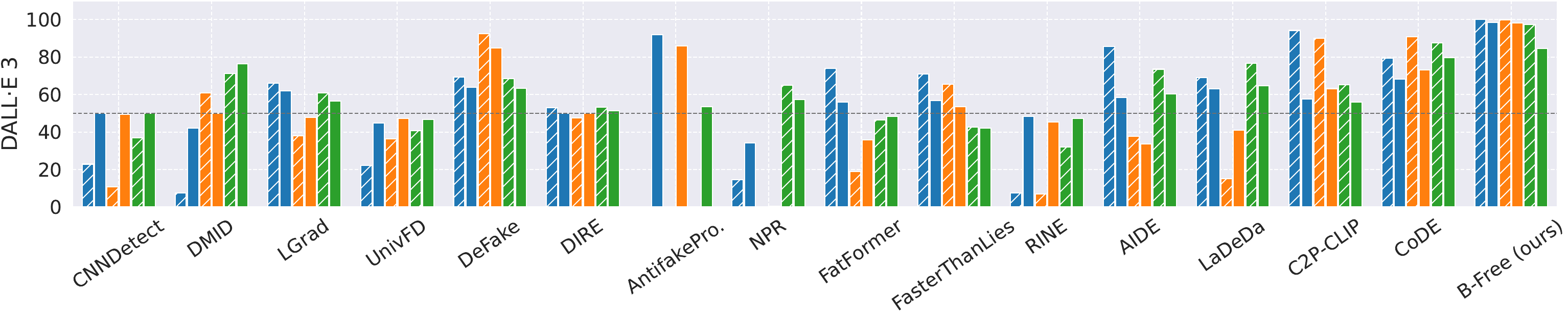} 
    \vspace{-1.4\baselineskip}
    \caption{SoTA performance evaluated in terms of AUC and balanced Accuracy on Midjourney, SDXL and DALL·E generators from different datasets.}
    \label{fig:same-gen-diff-data}
\end{figure*}

\vspace{2mm}
\noindent
\bb{FatFormer \cite{Liu2024forgery}.}
It adopts CLIP and introduces forgery-aware adapters to extract forensic traces from both space and frequency domains. The method proposes a language-guided alignment mechanism to supervise the process and ensure the association between image and text.

\vspace{2mm}
\noindent
\bb{FasterThanLies \cite{Lanzino2024faster}.}
The method employs a Binary Neural Network for features extraction phase and a linear classifier for detection.
Beyond the image, the model has two additional input channels: the Fast Fourier Transform magnitude and the Local Binary Pattern image.
We report results using the unfrozen BNext-M backbone.

\vspace{2mm}
\noindent
\bb{RINE \cite{Koutlis2024leveraging}.}
It uses features extracted from the intermediate blocks of a CLIP encoder and an additional trainable module to take into account the influence of each block on the final decision.

\vspace{2mm}
\noindent
\bb{AIDE \cite{Yan2024sanity}.}
It leverages hybrid features extracted from a ConvNeXt-based Open CLIP model and a CNN which is fed with patches filtered to remove semantic content and exploit low-level artifacts.

\vspace{2mm}
\noindent
\bb{LaDeDA \cite{Cavia2024real}.}
It is a patch-based classifier that leverages local image features. The image is split into multiple patches, for each patch a prediction is computed and then averaged to obtain the image-level prediction.

\vspace{2mm}
\noindent
\bb{C2P-CLIP \cite{Tan2024c2p}.} It uses the Low-Rank Adaptation (LoRA) strategy to fine-tune the image encoder of CLIP. Moreover, it relies on a contrastive learning strategy based on category prompts.

\vspace{2mm}
\noindent
\bb{CoDE \cite{Baraldi2024contrastive}.} 
CoDE trains a Vision Transformer using a contrastive loss similar to CLIP. However, while CLIP aims to learn features for text-image matching, CoDE aims at obtaining an embedding space where real and fake images are effectively separated. We report results using CoDE in combination with the linear classifier.

\section{Calibration Metrics}
\label{metrics}

Here we provide some more details about the calibration metrics used in the paper.
The binary Expected Calibration Error (ECE) is defined as:
\begin{equation}
\text{ECE} = \sum_{m=1}^M \frac{|B_m|}{N} \left| \text{prob}(B_m) - \text{pred}(B_m) \right|
\end{equation}
where $N$ is the number of samples of the test-set, $M$ is the number of bins,
and $B_m$ is the set of samples whose predictions fall into the $m$-th bin, with $|B_m|$ its cardinality.
$\text{prob}(B_m)$ and $\text{pred}(B_m)$ are the actual probability and the average predicted probability of the target class in that bin, respectively.
In case of unbalanced test-set, we weigh the contribution of each sample in the average to re-balance the relevance between two classes. We used $M=15$ bins.

The balanced Negative Log-Likelihood (NLL) is defined as:
\begin{equation}
\text{NLL} = - \frac{0.5}{|S_0|} \sum_{i \in S_0} \log p_i(0) - \frac{0.5}{|S_1|} \sum_{i \in S_1} \log p_i(1)
\end{equation}
where $S_0$ and $S_1$ are the set of samples of non-target and target class, respectively, while $p_i(0)$ and $p_i(1)$ are the predicted probabilities of the two classes for the $i$-th input sample.

\section{Additional Ablation}
\label{ablation}

In this Section we further investigate the influence of our training dataset.
We train our network on the very recent Diffusion-generated Deepfake Detection (D$^3$) dataset \cite{Baraldi2024contrastive}, of about 8M synthetic images (from 256 $\times$ 256 to 1024 $\times$ 1024) from the generators SD 1.4, SD 2.1, SDXL, and DeepFloyd IF. 
The images are generated using prompts taken from the description of the real source from LAION (text-driven generation).
From \cref{tab:results_abl_3}, we can notice that although the AUC is similar, there is a significant increase in terms of balanced accuracy (87.5 vs 95.2) and decrease in terms of both NLL (0.31 vs 0.14) and ECE (0.13 vs 0.04). This confirms that our training paradigm enables better calibration and improved generalization.

\begin{table*}[t!]
\centering
\scalebox{\scalefactortab}{\small
    \begin{tabular}{l  C{14mm}C{14mm}C{14mm}C{14mm} C{14mm}C{14mm}C{14mm}C{14mm} C{14mm}}
    \toprule
   \multirow{2}{*}{\b{AUC}$\uparrow$/\b{bAcc}$\uparrow$} & \multicolumn{9}{c}{\b{GenImage (unbiased)}}   \\  
   \cmidrule(lr){2-10}
   & BigGAN & VQDM & ADM &  GLIDE    & SD 1.4 & SD 1.5 & Midjourney & Wukong & AVG \\  
   \cmidrule(lr){1-1} \cmidrule(lr){2-9} \cmidrule(lr){10-10}
\ru CNNDetect         &    70.9  /    58.4  &    63.4  /    51.2  &    51.8  /    49.9  &    59.4  /    50.7  &    65.1  /    50.1  &    66.4  /    49.9  &    79.3  /    50.1  &    62.6  /    50.2  &    64.8  /    51.3  \\
\ru DMID              &    74.6  /    52.3  & \b{97.6} /    75.1  &    78.5  /    51.3  &    94.9  /    56.6  & \b{100.} / \b{99.9} & \b{100.} / \b{99.8} & \b{100.} / \b{97.4} & \b{100.} / \b{99.6} &    93.2  /    79.0  \\
\ru LGrad             &    18.7  /    28.9  &    23.9  /    30.8  &    24.6  /    30.5  &    22.2  /    30.0  &    50.0  /    49.8  &    49.2  /    49.1  &    50.5  /    50.6  &    47.6  /    46.9  &    35.8  /    39.6  \\
\ru UnivFD            &    96.7  /    86.1  &    94.8  /    79.7  &    85.2  /    64.4  &    88.8  /    63.9  &    78.7  /    55.5  &    78.1  /    56.6  &    74.0  /    54.2  &    86.9  /    63.7  &    85.4  /    65.5  \\
\ru DeFake            &    72.6  /    64.4  &    71.1  /    64.4  &    49.3  /    48.5  &    87.9  /    80.4  &    93.3  /    85.1  &    93.4  /    85.4  &    87.7  /    79.2  &    89.8  /    81.8  &    80.6  /    73.7  \\
\ru DIRE              &    26.6  /    46.9  &    35.0  /    47.7  &    25.3  /    46.7  &    29.9  /    47.0  &    41.7  /    47.3  &    39.8  /    47.3  &    38.0  /    47.5  &    45.4  /    47.7  &    35.2  /    47.3  \\
\ru AntifakePrompt    &    ~~~~-~~~  /    81.7  &    ~~~~-~~~  /    81.1  &    ~~~~-~~~  / \b{81.6} &    ~~~~-~~~  /    81.8  &    ~~~~-~~~  /    77.1  &    ~~~~-~~~  /    76.6  &    ~~~~-~~~  /    70.4  &    ~~~~-~~~  /    77.6  &    ~~~~-~~~  /    78.5  \\
\ru NPR               &    56.9  /    56.3  &    52.3  /    53.9  &    46.9  /    50.5  &    42.1  /    48.3  &    54.3  /    49.4  &    53.3  /    49.7  &    42.3  /    47.4  &    52.4  /    50.2  &    50.1  /    50.7  \\
\ru FatFormer         &    88.5  /    80.1  &    84.5  /    71.5  &    69.1  /    60.4  &    78.4  /    65.1  &    49.8  /    52.0  &    48.7  /    53.3  &    46.2  /    51.6  &    61.6  /    58.1  &    65.9  /    61.5  \\
\ru FasterThanLies    &    78.9  /    54.1  &    86.8  /    76.6  &    88.6  /    77.2  &    83.0  /    66.1  &    97.8  /    92.2  &    97.9  /    92.3  &    83.1  /    69.7  &    95.4  /    88.1  &    88.9  /    77.0  \\
\ru RINE              & \b{99.4} / \b{88.5} & \b{98.4} /    81.4  & \b{93.8} /    63.9  & \b{98.1} /    74.7  &    93.9  /    60.5  &    94.1  /    61.1  &    86.3  /    52.4  &    95.7  /    70.0  &    95.0  /    69.1  \\
\ru AIDE              &    73.1  /    50.7  &    78.0  /    51.0  &    61.2  /    50.1  &    80.4  /    52.3  &    98.2  /    74.5  &    98.5  /    75.9  &    88.1  /    57.4  &    95.9  /    69.3  &    84.2  /    60.2  \\
\ru LaDeDa            &    93.1  /    80.3  &    10.8  /    34.8  &    ~6.8  /    34.6  &    ~8.8  /    34.5  &    55.6  /    54.8  &    53.6  /    53.0  &    51.3  /    52.1  &    61.6  /    57.7  &    42.7  /    50.2  \\
\ru C2P-CLIP          &    97.2  /    87.5  &    92.2  /    74.1  &    86.7  /    71.3  &    93.6  /    74.8  &    94.4  /    80.5  &    94.3  /    79.1  &    76.3  /    55.9  &    93.1  /    81.0  &    91.0  /    75.5  \\
\ru CoDE              &    70.2  /    50.0  &    66.8  /    56.0  &    53.7  /    51.9  &    78.1  /    58.0  & \b{99.4} /    96.6  & \b{99.2} /    96.5  &    86.0  /    69.6  & \b{99.1} /    95.0  &    81.6  /    71.7  \\
\cmidrule(lr){1-1} \cmidrule(lr){2-9} \cmidrule(lr){10-10}
\ru B-Free (ours)     &    94.1  /    68.7  &    97.0  / \b{88.7} & \b{93.0} /    79.8  &    95.8  / \b{85.3} & \b{100.} /    98.8  & \b{100.} /    98.8  & \b{99.2} /    95.7  & \b{100.} / \b{99.0} & \b{97.4} / \b{89.3} \\
 \bottomrule
\end{tabular}
}
\vspace{-1.5mm}
\caption{Performance on each generator included in GenImage (unbiased) dataset in terms of AUC and balanced Accuracy.
Bold underlines the best performance for each column with a margin of 1\%.}
\label{tab:RES_genimage}
\end{table*}

\begin{table*}[t!]
\centering
\scalebox{\scalefactortab}{\small
    \begin{tabular}{l C{14mm}C{14mm}C{14mm}C{14mm}C{14mm}C{14mm}C{14mm}C{14mm}C{14mm}C{14mm}C{14mm}}
    \toprule
   \multirow{2}{*}{\b{AUC}$\uparrow$/\b{bAcc}$\uparrow$} & \multicolumn{11}{c}{\b{FakeBench}} \\
     \cmidrule(r){2-12}
     &  ProGAN & StyleGAN  & FuseDream  &  VQDM & GLIDE  & CogView2   & DALL·E 2 & DALL·E 3 & SD  & Midjourney & AVG \\  
     \cmidrule(r){1-1} \cmidrule(r){2-11} \cmidrule(lr){12-12}
\ru CNNDetect         & \b{100.} / \b{99.7} &    98.3  /    75.1  &    94.8  /    61.1  &    62.9  /    51.9  &    62.6  /    50.6  &    64.9  /    49.7  &    56.1  /    49.7  &    58.6  /    49.7  &    57.2  /    49.7  &    62.0  /    49.9  &    71.7  /    58.7  \\
\ru DMID              &    61.0  /    51.1  &    80.1  /    52.1  &    93.1  /    52.4  &    97.8  /    79.7  &    94.0  /    63.2  & \b{100.} / \b{99.7} &    94.9  /    55.1  &    96.7  /    88.9  & \b{100.} / \b{99.1} & \b{97.3} /    90.7  &    91.5  /    73.2  \\
\ru LGrad             &    96.8  /    77.1  &    82.3  /    72.9  &    18.9  /    28.4  &    75.2  /    68.6  &    41.8  /    43.9  &    23.7  /    33.6  &    10.9  /    27.6  &    30.6  /    35.6  &    24.7  /    34.1  &    76.1  /    67.3  &    48.1  /    48.9  \\
\ru UnivFD            & \b{99.9} /    98.6  &    96.0  /    83.4  & \b{99.2} / \b{96.3} &    94.6  /    77.3  &    86.5  /    62.8  &    84.7  /    63.1  &    88.0  /    65.9  &    69.6  /    55.8  &    76.8  /    56.4  &    65.5  /    55.6  &    86.1  /    71.5  \\
\ru DeFake            &    63.7  /    58.1  &    73.7  /    66.7  &    53.8  /    51.0  &    69.8  /    64.5  &    81.6  /    74.2  &    84.7  /    77.2  &    83.6  /    76.5  &    81.7  /    74.5  &    86.4  /    77.3  &    78.7  /    70.5  &    75.8  /    69.0  \\
\ru DIRE              &    90.4  /    89.5  &    56.6  /    55.4  &    23.7  /    40.0  &    91.3  /    89.2  &    53.2  /    63.7  &    36.7  /    41.0  &    44.2  /    43.0  &    76.6  /    74.5  &    47.7  /    49.7  &    83.4  /    81.2  &    60.4  /    62.7  \\
\ru AntifakePrompt    &    ~~~~-~~~  /    79.0  &    ~~~~-~~~  /    78.0  &    ~~~~-~~~  /    78.6  &    ~~~~-~~~  /    77.0  &    ~~~~-~~~  /    78.8  &    ~~~~-~~~  /    75.8  &    ~~~~-~~~  /    73.4  &    ~~~~-~~~  /    74.0  &    ~~~~-~~~  /    71.6  &    ~~~~-~~~  /    76.1  &    ~~~~-~~~  /    76.2  \\
\ru NPR               & \b{99.5} /    92.4  &    78.1  /    68.1  &    48.7  /    42.7  &    93.4  / \b{90.9} &    67.0  /    65.2  &    50.3  /    42.7  &    41.7  /    42.9  &    46.3  /    44.6  &    57.5  /    51.4  &    89.0  /    84.6  &    67.1  /    62.5  \\
\ru FatFormer         & \b{100.} /    97.6  & \b{99.3} / \b{97.1} &    90.7  /    81.8  &    96.8  /    88.5  &    74.2  /    69.0  &    47.1  /    53.3  &    45.3  /    48.1  &    52.4  /    49.5  &    50.4  /    51.0  &    79.6  /    64.6  &    73.6  /    70.0  \\
\ru FasterThanLies    &    87.0  /    80.2  &    72.4  /    57.7  &    85.7  /    75.4  &    54.7  /    45.9  &    76.9  /    62.5  &    96.0  /    87.7  &    92.9  /    85.7  &    73.6  /    60.6  &    93.6  /    84.5  &    67.6  /    57.7  &    80.0  /    69.8  \\
\ru RINE              & \b{100.} / \b{99.6} & \b{99.3} /    95.1  & \b{99.8} / \b{96.6} & \b{98.8} /    88.6  & \b{95.4} /    70.2  &    86.7  /    59.2  &    93.0  /    60.9  &    75.1  /    52.6  &    85.5  /    55.9  &    82.2  /    61.1  &    91.6  /    74.0  \\
\ru AIDE              &    89.4  /    64.3  &    89.4  /    70.0  &    71.7  /    47.3  &    90.7  /    78.1  &    79.7  /    68.3  &    85.5  /    60.0  &    84.1  /    52.6  &    88.0  /    61.9  &    86.0  /    64.6  &    88.0  /    71.5  &    85.2  /    63.9  \\
\ru LaDeDa            &    98.0  /    82.5  &    94.5  /    82.5  &    37.2  /    40.5  &    85.8  /    81.5  &    52.6  /    57.3  &    39.4  /    41.6  &    36.4  /    35.3  &    49.9  /    45.1  &    45.3  /    46.1  &    90.7  /    78.1  &    63.0  /    59.1  \\
\ru C2P-CLIP          & \b{100.} / \b{99.5} & \b{99.4} / \b{98.0} &    98.2  /    93.0  &    97.1  /    86.7  &    91.9  /    76.4  &    67.3  /    61.7  &    72.6  /    56.9  &    74.7  /    55.5  &    74.9  /    59.9  &    88.1  /    58.0  &    86.4  /    74.6  \\
\ru CoDE              &    64.3  /    52.5  &    53.0  /    49.5  &    73.4  /    56.3  &    78.4  /    61.7  &    91.6  /    78.0  &    97.7  /    93.7  &    93.8  /    82.8  &    95.8  /    89.2  & \b{99.5} /    96.2  &    89.7  /    76.7  &    83.7  /    73.7  \\
\cmidrule(r){1-1} \cmidrule(r){2-11} \cmidrule(lr){12-12} 
\ru B-Free (ours)     & \b{99.3} /    96.4  &    97.7  /    88.5  & \b{99.3} /    95.2  &    96.5  /    86.5  & \b{95.5} / \b{87.7} & \b{100.} / \b{98.9} & \b{98.7} / \b{94.9} & \b{99.9} / \b{98.7} & \b{100.} / \b{98.7} & \b{98.0} / \b{94.5} & \b{98.5} / \b{94.0} \\
 \bottomrule
\end{tabular}
}
\vspace{-1.5mm}
\caption{Performance on each generator included in FakeBench dataset in terms of AUC and balanced Accuracy. Bold underlines the best performance for each column with a margin of 1\%.}
\label{tab:RES_fakebench}
\end{table*}

\section{Additional Generalization Analysis}
\label{generalization}

In this Section we conduct further experiments which confirm that our method can generalize better than other methods and obtain less biased results. We also detail results and show the performance on each synthetic generator for GenImage and FakeBench datasets.

\vspace{2mm}
\noindent
\bb{Evaluation on same generators from different datasets.}
Here we further expand on the analysis conducted in Fig.~\ref{fig:biases}, where we have shown that some detectors achieve different performance on the same generator when the images are taken from two different datasets. This puzzling behavior suggests the possibility that these methods rely on subtle dataset biases besides true traces left by the synthetic generator. In Fig.~\ref{fig:same-gen-diff-data} we extend this analysis to all SoTA methods described in Sec.~\ref{methods}. More specifically, we analyze the performance in terms of AUC and balanced accuracy over three synthetic generators: Midjourney, SDXL and DALL-E 3 that come from three different datasets PolarDiffShield \cite{Li2024masksim}, Synthbuster \cite{Bammey2023synthbuster} and FakeInversion \cite{Cazenavette2024fakeinversion}. 
As said before, for several methods the performance is not consistent on the same generator and can vary even by 20\% from one dataset to another. In addition, for some methods the AUC is around 50\%, which corresponds to random choice, or even below 50\% which means that the detector tends to invert the labels between real and fake.

\vspace{2mm}
\noindent
\bb{Evaluation on different synthetic generators.}
We conduct a more detailed analysis of the results on GenImage (unbiased), where fake images have been subjected to JPEG compression, similar to real images, to prevent detectors from exploiting compression artifacts. We also consider FakeBench \cite{Li2024fakebench}, that consists of 3,000 real and 3,000 fake images generated by 10 different models. These datasets include both GAN and Diffusion-based synthetic images, which allows us to better understand the ability of our approach to generalize to different architectures. Results are presented in Tab.~\ref{tab:RES_genimage} and Tab.~\ref{tab:RES_fakebench}. We note that our approach  obtains very good results consistently across almost all generators, while other methods, such as DMID, UnivFD, RINE, FasterThanLies, and FatFormer, perform very well in terms of AUC only on certain generators. In addition, for our method the gap between AUC and balanced accuracy is reduced which ensures more reliable results.

Finally, we generated a set of 2,000 images with two autoregressive models \cite{Li2024autoregressive, Yu2024randomized}. 
whose architecture is significantly different from the one used to generate the training data. 
On both models we obtain good results with an average of 99\% in terms of AUC and 94.2\% in terms of Accuracy. This is probably due to the similarity between the tokenizer used in these models and the latent embedders of Stable Diffusion models. 

\begin{figure}[h!]
    \centering
    \vspace{5.3mm}
    \includegraphics[width=0.32\linewidth, page=1, trim=12 0 0 0,clip]{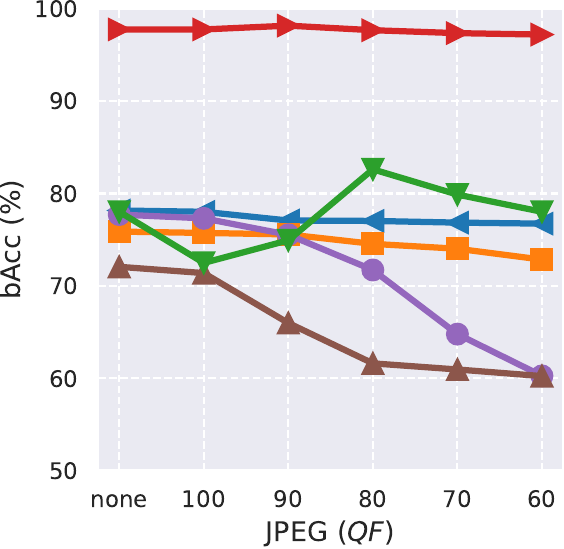}
    \includegraphics[width=0.32\linewidth, page=2, trim=12 0 0 0,clip]{figures/results_robustness_supp_bAcc_SOTA.pdf}
    \includegraphics[width=0.32\linewidth, page=3, trim=12 0 0 0,clip]{figures/results_robustness_supp_bAcc_SOTA.pdf} 
    \\[1mm]
    \includegraphics[width=0.99\linewidth, page=4]{figures/results_robustness_supp_bAcc_SOTA.pdf}\vspace{1.2mm}
    \caption{Robustness analysis in terms of balanced Accuracy carried out on nine generators of Synthbuster under three different post-processing operations: JPEG compression, resizing and blurring. The five best performing SoTA detectors on Synthbuster have been included in the analysis.}
    \label{fig:robustness_appendix_DINO}
\end{figure}

\section{Additional Robustness Analysis}
\label{robustness}

We repeat the experiment reported in Figure \ref{fig:robustness} to test robustness under various operations and carry out comparisons with SOTA methods.
Results are shown in Fig.~\ref{fig:robustness_appendix_DINO} under three post-processing operations: JPEG compression, resizing, and blurring. We can observe that our approach is more robust by a large margin compared with other methods and can ensure a balanced accuracy that is always above 80\% even in the most challenging scenario.

Finally we further investigate robustness performance on FakeInversion \cite{Cazenavette2024fakeinversion}, where the real images have been retrieved from the web. To better understand the effect of compression and resizing we compare the performance when applying such operations. In particular, to simulate the upload on social networks, we resize with a scale factor randomly sampled between 0.7 and 1, and compress with a JPEG quality factor between 70 and 100. In Tab.~\ref{tab:RES_fi} results show that the performance on such datase drops substantially, except for DMID and our method, though our approach has better calibration metrics.

\begin{table}[t]
\centering
\scalebox{\scalefactortab}{\small
\begin{tabular}{l C{15mm}C{15mm}C{15mm}C{15mm}}
	\toprule
    & \multicolumn{2}{c}{\b{Original}} & \multicolumn{2}{c}{\b{Social network simulation}} \\
    \cmidrule(r){2-3} \cmidrule(r){4-5}
	&  AUC$\uparrow$/bAcc$\uparrow$ & NLL$\downarrow$/ECE$\downarrow$  &  AUC$\uparrow$/bAcc$\uparrow$ & NLL$\downarrow$/ECE$\downarrow$   \\  
   \cmidrule(r){1-1}   \cmidrule(r){2-3} \cmidrule(r){4-5}
\ru CNNDetect                  &    54.3  /    50.9  &    7.94  /    .488  &    51.8  /    50.1  &    8.73  /    .498\\
\ru DMID                       &    97.3  / \b{96.1} & \b{0.25} / \b{.041} &    94.4  /    81.3  &    0.55  /    .182\\
\ru LGrad                      &    84.4  /    77.2  &    2.27  /    .200  &    60.2  /    55.4  &    7.59  /    .426\\
\ru UnivFD                     &    54.9  /    52.8  &    2.19  /    .391  &    49.7  /    50.0  &    2.57  /    .434\\
\ru DeFake                     &    69.8  /    63.3  &    0.95  /    .225  &    69.3  /    62.7  &    0.98  /    .236\\
\ru DIRE                       &    53.3  /    51.8  &    13.4  /    .360  &    55.5  /    51.7  &    13.4  /    .358\\
\ru AntifakePrompt             &    ~~~-~~~  /    53.9  &     ~~~-~~~  /    ~~~-~~~   &    ~~~-~~~  /    54.8  &     ~~~-~~~  /    ~~~-~~~ \\
\ru NPR                        &    91.6  /    87.0  &    4.96  /    .123  &    43.3  /    49.9  &    27.1  /    .501\\
\ru FatFormer                  &    68.2  /    59.7  &    3.45  /    .386  &    48.7  /    50.4  &    5.44  /    .490\\
\ru FasterThanLies             &    49.7  /    48.6  &    3.64  /    .476  &    51.0  /    49.9  &    3.13  /    .458\\
\ru RINE                       &    69.6  /    63.6  &    4.84  /    .319  &    62.4  /    52.9  &    6.28  /    .437\\
\ru AIDE                       &    85.5  /    76.9  &    0.54  /    .137  &    67.3  /    56.4  &    0.93  /    .276\\
\ru LaDeDa                     &    91.7  /    84.7  &    3.03  /    .129  &    51.9  /    53.1  &    24.8  /    .454\\
\ru C2P-CLIP                   &    74.1  /    59.6  &    0.82  /    .260  &    71.9  /    59.0  &    0.89  /    .284\\
\ru CoDE                       &    87.5  /    78.7  &    0.74  /    .143  &    82.5  /    74.4  &    0.89  /    .173\\
\cmidrule(r){1-1}   \cmidrule(r){2-3} \cmidrule(r){4-5}
\ru B-Free (ours)   & \b{99.3} /    86.3  &    0.32  /    .144  & \b{98.5} /  \b{86.4}  &  \b{0.32} / \b{.131}\\
	\bottomrule
\end{tabular}
}
\vspace{-1.5mm}
\caption{Performance on FakeInversion dataset. We show results on the original dataset and on a post-processed version, to simulate the upload on social networks.}
\label{tab:RES_fi}
\end{table}

\end{appendix}

\balance
{
    \small
    \bibliographystyle{ieeenat_fullname}
    \bibliography{main}
}

\end{document}